\documentclass{article}

    \PassOptionsToPackage{numbers, compress}{natbib}

    \usepackage[preprint]{neurips_2024}

\usepackage[utf8]{inputenc} 
\usepackage[T1]{fontenc}    
\usepackage{hyperref}       
\usepackage{url}            
\usepackage{booktabs}       
\usepackage{amsfonts}       
\usepackage{nicefrac}       
\usepackage{microtype}      
\usepackage{xcolor}         
\usepackage{amsmath,amssymb}
\usepackage{algorithm}
\usepackage{algpseudocode}
\usepackage{graphicx}

\newcommand*{\dif}{\mathop{}\!\mathrm{d}}

\title{VAE-Var: Variational-Autoencoder-Enhanced Variational Assimilation}

\author{%
  Yi Xiao \\
  Tsinghua University \\
  Shanghai Artificial Intelligence Laboratory\\
  \texttt{xiaoyi200018@gmail.com} \\
  \And
  Qilong Jia \\
  Tsinghua University \\
  \texttt{jql23@mails.tsinghua.edu.cn} \\
  \AND
  Wei Xue \\
  Tsinghua University \\
  \texttt{xuewei@tsinghua.edu.cn} \\
  \And
  Lei Bai \\
  Shanghai Artificial Intelligence Laboratory\\
  \texttt{bailei@pjlab.org.cn} \\
}

\begin{document}

\maketitle

\begin{abstract}
Data assimilation refers to a set of algorithms designed to compute the optimal estimate of a system's state by refining the prior prediction (known as background states) using observed data. Variational assimilation methods rely on the maximum likelihood approach to formulate a variational cost, with the optimal state estimate derived by minimizing this cost. Although traditional variational methods have achieved great success and have been widely used in many numerical weather prediction centers, they generally assume Gaussian errors in the background states, which limits the accuracy of these algorithms due to the inherent inaccuracies of this assumption. 
In this paper, we introduce VAE-Var, a novel variational algorithm that leverages a variational autoencoder (VAE) to model a non-Gaussian estimate of the background error distribution. We theoretically derive the variational cost under the VAE estimation and present the general formulation of VAE-Var; we implement VAE-Var on low-dimensional chaotic systems and demonstrate through experimental results that VAE-Var consistently outperforms traditional variational assimilation methods in terms of accuracy across various observational settings.

\end{abstract}

\section{Introduction}

Data assimilation is a statistical technique used to produce accurate estimate (known as the analysis states) of a physical system's states by blending prior predictions (known as the background states) with observational data. This process is crucial for deriving initial states in many fields, particularly in numerical weather forecasting~\citep{bauer2015quiet, kalnay2003atmospheric, carrassi2018data}.

For over two decades, the variational assimilation algorithm has been the mainstay of data assimilation in numerical weather prediction centers~\citep{rabier1998extended, rabier2000ecmwf}. Variational assimilation stands for a group of algorithms, in which a posterior likelihood of the physical states to be solved is calculated according to Bayes' theorem, and the negative logarithm of the posterior likelihood is known as the variational cost. The analysis states are obtained by minimizing the variational cost via gradient descent~\citep{asch2016data, navon2009data, talagrand1987variational}. 

The primary challenge of the variational assimilation lies in accurately approximating the background error distribution, which directly governs how the observational data should be utilized to refine the system's states~\citep{descombes2015generalized}. Consequently, significant efforts in data assimilation algorithm development are directed towards enhancing background error distribution estimation. For instance, in the WRF-DA system, the background error covariance is decomposed into horizontal, vertical, and cross-variable correlations to effectively represent various aspects of background error structures~\citep{huang2009four}; ECMWF utilizes the EDA (Ensembles of Data Assimilations) method to better capture variations in the background error distribution~\citep{bonavita2012use}; additionally, in FengWu-4DVar, spherical harmonic transformations are employed to accurately represent horizontal correlations on the sphere~\citep{xiao2023fengwu}. 

Despite the success of current progress in background error estimation, almost all works rely on the assumption that the error distribution of the background state is Gaussian~\citep{barker2004three, bannister2008review, descombes2015generalized, tr2006accounting}. However, since the background state is obtained by integrating the numerical model, and the non-linearity of the numerical model is strong~\citep{strogatz2018nonlinear}, this usually results in a non-Gaussian distribution of the background states. Therefore, the Gaussian distribution assumption potentially hinders the accuracy of Bayesian estimates of the analysis state distribution and further limits the final assimilation accuracy. 

Generative neural networks, known for their capability to learn complex distributions, have seen rapid advancements in various fields~\citep{ho2020denoising, lugmayr2022repaint, chung2022diffusion, kingma2013auto, girin2020dynamical}. Nevertheless, to the best of our knowledge, no work has leveraged these networks to capture the non-Gaussian characteristics of background error for improving variational assimilation algorithms. While learning the background error distribution with these powerful neural network tools is generally feasible, the challenge lies in integrating neural network-based distribution formulations into the variational cost and simplifying this cost, which may include complex integrals, to make it optimizable. In this work, we address this challenge by (1) utilizing a variational autoencoder (VAE), which offers a more concise distribution formulation than diffusion models, to provide a parametric estimate of the background error and (2) introducing several reasonable assumptions about the VAE to formulate an optimizable variational cost.


\paragraph{Contribution} We propose a novel variational assimilation algorithm, named VAE-Var. This algorithm leverages VAE~\cite{kingma2013auto} to obtain a non-Gaussian parametric estimate of the background error distribution, offering a more accurate and reliable estimate compared to the Gaussian assumption. We outline the general formulation of the VAE-Var variational cost and provide an implementation strategy on low-dimensional dynamic systems.
The experimental results on two classical chaotic systems demonstrate that our proposed algorithm consistently outperforms traditional variational assimilation in terms of assimilation accuracy across different settings, including both 3DVar and 4DVar observation scenarios with linear or nonlinear observation operators. 

\section{Preliminaries}
\label{sec:preliminaries}

\paragraph{Variational Assimilation} 

Variational assimilation seeks to determine the probability density function $p(\mathbf{x}|\mathbf{y})$ of the physical state's distribution at a specific time, given known observational conditions $\bf y$, and calculate the optimal estimate of the physical state (referred to as the analysis state $\mathbf{x}_a$) by maximizing this probability density function, that is, $\mathbf{x}_a = \arg \max_{\mathbf{x}} p(\mathbf{x}|\mathbf{y})$.

According to the Bayes' Theorem,
\begin{equation}
    \arg \max_{\mathbf{x}} p(\mathbf{x}|\mathbf{y}) = \arg \max_{\mathbf{x}} \frac{p(\mathbf{y}|\mathbf{x}) p(\mathbf{x})}{p(\mathbf{y})} = \arg \max_{\mathbf{x}} p(\mathbf{y}|\mathbf{x}) p(\mathbf{x}).
\end{equation}
We define the variational cost as:
\begin{equation}
    \mathcal{L}(\mathbf{x}) = -\log p(\mathbf{y}|\mathbf{x}) p(\mathbf{x}) = - \log p(\mathbf{y}|\mathbf{x}) - \log p(\mathbf{x}).
\end{equation}
Maximizing the probability density function is equivalent to minimizing the variational cost. This cost comprises two terms: an observation term, $\mathcal{L}_o(\mathbf{x}, \mathbf{y}) = - \log p(\mathbf{y}|\mathbf{x})$ and a prior term, $\mathcal{L}_p(\mathbf{x}) = - \log p(\mathbf{x})$. 
The prior term is also referred to as the background term, denoted as $\mathcal{L}_b(\mathbf{x}, \mathbf{x}_b)$. In the context of data assimilation, the background term captures the discrepancy between the current physical state $\mathbf{x}$ and the predicted state from a previous time step, known as the background state $\mathbf{x}_b$. 

\paragraph{Observation Term} The physical states of a dynamical system can be observed through instruments, where the conversion function from the physical state to the observation is denoted by $\mathcal{H}$. Also, since instrument measurements introduce Gaussian errors, we can assume that $\mathbf{y} | \mathbf{x} \sim \mathcal{N}(\mathcal{H}(\mathbf{x}), \mathbf{R})$, where $\mathbf{R}$ corresponds to the covariance of the observation error. The observation term of the variational cost can be formulated as follows for three-dimensional variational assimilation (3DVar):
\begin{equation}
\label{eq-3dvar-obs}
    \mathcal{L}_o(\mathbf{x}, \mathbf{y}) = \frac12 (\mathbf{y} - \mathcal{H}(\mathbf{x}))^\mathrm{T} \mathbf{R}^{-1} (\mathbf{y} - \mathcal{H}(\mathbf{x})).
\end{equation}
In operational weather forecasting systems, both current and subsequent time observations are utilized to assimilate the physical state at the current moment, known as four-dimensional variational assimilation (4DVar). Let $\{\mathbf{y}_i\}_{i=0}^{N-1}$ represent the observations at times $t_0, ..., t_{N-1}$ and $\mathcal{M}_{0\to n}$ denote the forecasting model from time $t_0$ to $t_n$. The observation term of 4DVar can be expressed as:
\begin{equation}
\label{eq-4dvar-obs}
    \mathcal{L}_o(\mathbf{x}, \mathbf{y}_0, ..., \mathbf{y}_{N-1}) = \frac12 \sum_{n=0}^{N-1} (\mathbf{y} - \mathcal{H}(\mathcal{M}_{0\to n}(\mathbf{x})))^\mathrm{T} \mathbf{R}^{-1} (\mathbf{y} - \mathcal{H}(\mathcal{M}_{0\to n}(\mathbf{x}))).
\end{equation}
Additional details about 4DVar are provided in Appendix~\ref{sec-4dvar-detail}.

\begin{figure}
    \centering
    \includegraphics[width=1.0\linewidth]{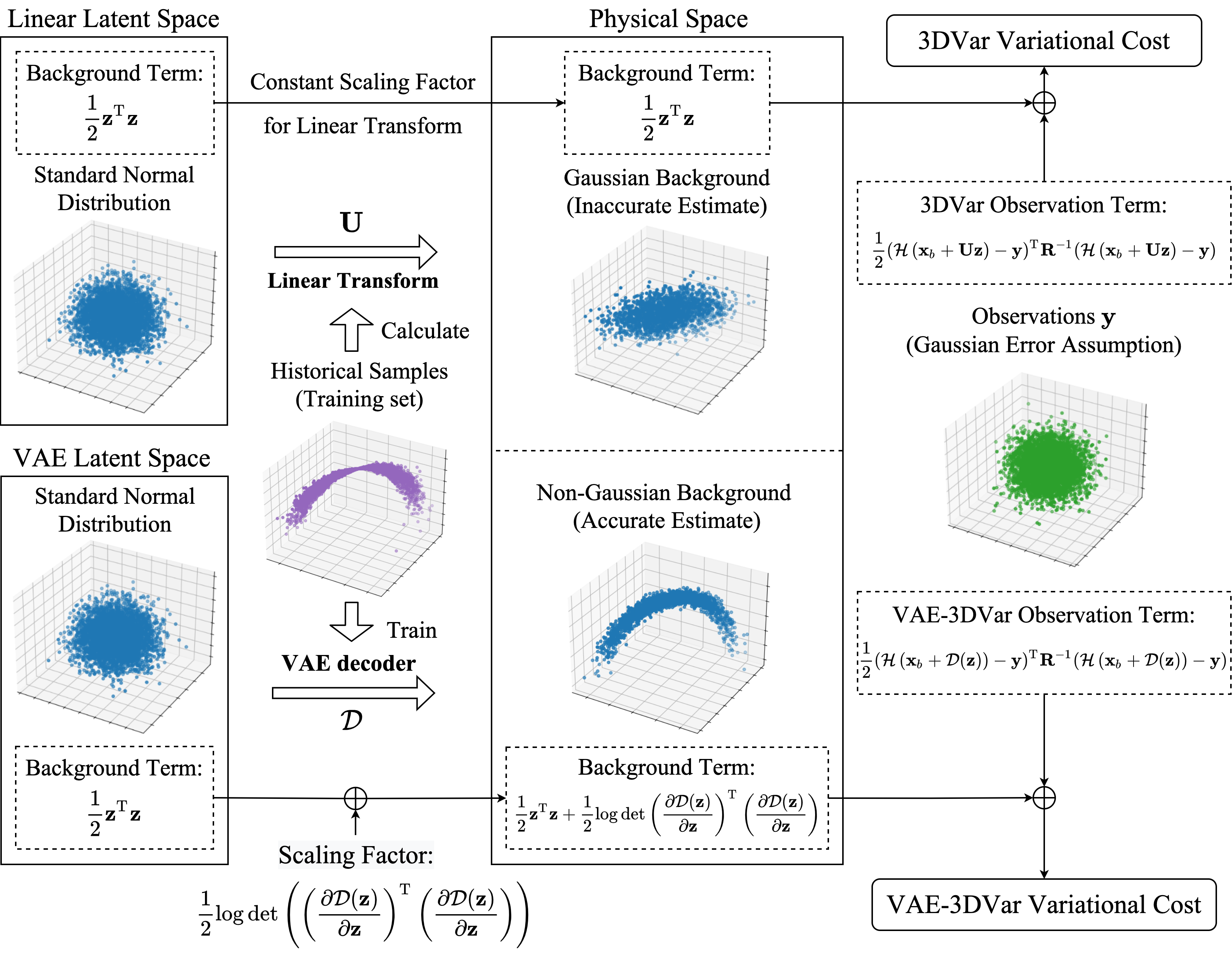}
    \caption{Comparison between the traditional 3DVar algorithm (upper part) and our proposed VAE-3DVar algorithm (lower part). The blue dots correspond to the distribution of the background state (in both the latent space and the physical space); the green dots correspond to the observation distribution; the purple dots correspond to the training samples generated with the NMC method.}
    \label{fig-framework}
\end{figure}

\paragraph{Background Term} 
As previously mentioned, the background state $\mathbf{x}_b$ is derived by integrating a numerical model from a previous state. In most cases, the numerical model typically cannot perfectly capture all dynamical information and is subject to inherent errors. Traditional variational assimilation algorithms assume these errors follow a Gaussian distribution $\mathbf{x} - \mathbf{x}_b \sim \mathcal{N}(0, \mathbf{B})$, where $\mathbf{B}$ is the background error covariance matrix estimated from historical error samples using methods like the National Meteorological Center (NMC) technique~\citep{descombes2015generalized}, which will be discussed in the subsequent section. The background term of the variational cost can thus be formulated as a quadratic function:
\begin{equation}
    \mathcal{L}_b^{trad}(\mathbf{x}, \mathbf{x}_b) = \frac12 (\mathbf{x} - \mathbf{x}_b)^\mathrm{T} \mathbf{B}^{-1} (\mathbf{x} - \mathbf{x}_b).
\end{equation}
To avoid calculating the inverse of $\mathbf{B}$ in the traditional variational assimilation, a linear variable transformation $\mathbf{x} = \mathbf{U}\mathbf{z} + \mathbf{x}_b$ is commonly used, where $\mathbf{U}$ satisfies $\mathbf{B} = \mathbf{U}^\mathrm{T}\mathbf{U}$. The background term can then be expressed as:
\begin{equation}
    \tilde{\mathcal{L}}_b^{trad}(\mathbf{z}) = \mathcal{L}_b^{trad}(\mathbf{U}\mathbf{z} + \mathbf{x}_b, \mathbf{x}_b) = \frac12 \mathbf{z}^\mathrm{T}\mathbf{z}.
\end{equation}
Similarly, the observation term can also be expressed as a function of $\mathbf{z}$. As for 3DVar, $\tilde{\mathcal{L}}^{trad}_o(\mathbf{z}) = \mathcal{L}_o(\mathbf{U}\mathbf{z} + \mathbf{x}_b, \mathbf{y})$; for 4DVar, $\tilde{\mathcal{L}}^{trad}_o(\mathbf{z}) = \mathcal{L}_o(\mathbf{U}\mathbf{z} + \mathbf{x}_b, \mathbf{y}_0, ..., \mathbf{y}_{N-1})$. The calculation of the background term, the observation term and the variational cost $\left(\tilde{\mathcal{L}}^{trad}(\mathbf{z}) = \tilde{\mathcal{L}}^{trad}_o(\mathbf{z}) + \tilde{\mathcal{L}}^{trad}_b(\mathbf{z})\right)$ in the traditional 3DVar algorithm is visualized in the upper part of Figure~\ref{fig-framework}. By minimizing the total variational cost $\mathbf{z}^\star = \arg \min_{\mathbf{z}} \tilde{\mathcal{L}}^{trad}(\mathbf{z})$, we obtain the optimal analysis state as $\mathbf{x}_a = \mathbf{U}\mathbf{z}^\star + \mathbf{x}_b$.

\section{VAE-Var}

\subsection{Background Error Estimation Based on VAE}
\label{sec-vae-bgerr}

While assuming a Gaussian distribution for the background state error can simplify problem-solving, this assumption is often unrealistic or at least inaccurate. In data assimilation, the background error mostly arises from model inaccuracies. Let $\mathcal{M}$ represent the numerical integration model used for prediction and $\mathcal{M}^{gt}$ denote the ground truth integration model. The distribution of the error in $\mathbf{x}_b$ can be estimated by constructing samples from $\mathcal{M}^{gt}(\mathbf{x}) - \mathcal{M}(\mathbf{x})$, where $\mathbf{x}$ is randomly sampled from physical states. Due to the complex relationship between $\mathcal{M}$ and $\mathcal{M}^{gt}$, these samples $\mathcal{M}^{gt}(\mathbf{x}) - \mathcal{M}(\mathbf{x})$ generally do not follow a Gaussian distribution.

We use the Lorenz 63 system to provide an illustrative example. This system involves three parameters: $\sigma$, $\rho$, $\beta$, as shown in Equation~\ref{eq-Lorenz-63}. We create dynamical models using two different sets of parameter values: $\sigma=10, \rho=28, \beta=\frac{8}{3}$ (for ground truth model $\mathcal{M}^{gt}$) and $\sigma=10, \rho=29, \beta=\frac{8}{3}$ (for prediction model $\mathcal{M}$). By numerically integrating these models from randomly-chosen identical initial states and calculating the difference between their outputs, we generate a set of error samples, as depicted with the purple dots in Figure~\ref{fig-framework}. The error distribution is observed to be distinctly non-Gaussian and exhibits a non-convex structure.

In our work, we aim to model the non-Gaussian feature of the background error $\delta = \mathbf{x} - \mathbf{x}_b$ with the help of variational autoencoder (VAE), which is a powerful tool for learning parametric distributions from data sets. According to the theory of variational Bayesian inference, we assume that the distribution of $\delta$ can be modeled by a conditional distribution $p_{\delta|\mathbf{z}}(\delta|\mathbf{z})$, where $\mathbf{z}$ is a latent variable with a prior distribution $p_\mathbf{z}(\mathbf{z})$ that follows a standard normal distribution. That is,
\begin{equation}
    p_\delta(\delta) = \int_{\mathbf{z}} p_{\delta|\mathbf{z}}(\delta|\mathbf{z}) p_\mathbf{z}(\mathbf{z}) \dif \mathbf{z}.
\end{equation}
By training a variational autoencoder, $p_{\delta|\mathbf{z}}(\delta|\mathbf{z})$ can be estimated. Specifically, denote $\mathcal{D}$ the decoder of the VAE, then $\delta|\mathbf{z} \sim \mathcal{N}(\mathcal{D}(\mathbf{z}), \sigma_0^2I_n)$, where $\sigma_0$ corresponds to the hyper-parameter during the training of VAE and $n$ corresponds to the dimension of the dynamical system.

The background term can be according expressed as
\begin{equation}
\begin{aligned}
    &\mathcal{L}_b^{vae}(\mathbf{x}, \mathbf{x}_b) = -\log \int_{\mathbf{z}} p_{\delta|\mathbf{z}}(\mathbf{x} - \mathbf{x}_b|\mathbf{z}) p_\mathbf{z}(\mathbf{z}) \dif \mathbf{z} \\
    =& -\log \int_{\mathbf{z}} \sigma_0^{-n} \exp{\left(-(\mathbf{x}-\mathbf{x}_b-\mathcal{D}(\mathbf{z}))^\mathrm{T}(\mathbf{x}-\mathbf{x}_b-\mathcal{D}(\mathbf{z}))/(2\sigma_0^2)\right)} \exp{\left(-\mathbf{z}^\mathrm{T}\mathbf{z}/2\right)} \dif \mathbf{z} + Constant.
\end{aligned}
\end{equation}

The total variational cost would be too complicated to optimize if this background term is directly adopted, but we can make the following three assumptions to make the problem tractable.

The first is assumption that we assume $\sigma_0$ to be close to zero ($\sigma_0 \to 0$). This parameter acts as a regularization term to manage sample errors during the training of a VAE; during the inference process, $\sigma_0$ is essentially set to zero, meaning that no additional noise is introduced after decoding the hidden state. This assumption is consistent with the inference process of a VAE.

Our second assumption is that the decoder $\mathcal{D}: \mathbb{R}^n \to \Omega \subset \mathbb{R}^N$ is bijective. This requires the encoder to learn perfect dimension reduction operations from high-dimensional manifolds. While it's challenging to fully validate this assumption in real-world scenarios, we can generally expect that this bijective property holds within local regions if a VAE is well-trained. 

Thirdly, we additionally require that the decoder satisfy certain non-singularity properties. Specifically, we demand the existence of a positive number $C$ such that for any $\mathbf{z} \in \mathbb{R}^n$, $\det\left(\left(\frac{\partial \mathcal{D}(\mathbf{z})}{\partial \mathbf{z}}\right)^\mathrm{T}\left(\frac{\partial \mathcal{D}(\mathbf{z})}{\partial \mathbf{z}}\right)\right) > C$ holds.

Based on the assumptions above, we can transform the objective function for the physical state $\mathbf{x}$ into an objective function for the latent state $\mathbf{z}$ using the transformation function $\mathbf{x} = \mathcal{D}(\mathbf{z}) + \mathbf{x}_b$. The background term with regard to the latent state can be formulated as:
\begin{equation}
\label{eq-vae-loss}
    \tilde{\mathcal{L}}_b^{vae}(\mathbf{z}) = \mathcal{L}_b^{vae}(\mathbf{x}, \mathbf{x}_b) = \frac12 \mathbf{z}^\mathrm{T}\mathbf{z} + \frac12 \log \det\left(\left(\frac{\partial \mathcal{D}(\mathbf{z})}{\partial \mathbf{z}}\right)^\mathrm{T}\left(\frac{\partial \mathcal{D}(\mathbf{z})}{\partial \mathbf{z}}\right)\right),
\end{equation}
where $\frac{\partial \mathcal{D}(\mathbf{z})}{\partial \mathbf{z}}$ corresponds to the Jacobian matrix of $\mathcal{D}$ and $\det(\cdot)$ corresponds to the determinant. The detailed derivation is provided in Appendix~\ref{sec-bg-term-derivation} for reference. Equation~\ref{eq-vae-loss} has very clear physical meanings: the first term $\frac12 \mathbf{z}^\mathrm{T}\mathbf{z}$ corresponds to the background term (also called the regularization term) of the latent states $\mathbf{z}$ in the latent space, denoted as $\mathcal{L}_{reg}$; the second term represents the scaling determinant that converts the regularization term in the latent space into physical space, denoted as $\mathcal{L}_{det}$. The formulation of VAE-Var closely resembles that of the traditional variational algorithm, with an extra addition of a scaling determinant term. This term is essential because VAE-Var introduces a nonlinear relationship between the latent space and the original physical space, whereas in traditional 3DVar, this relationship is linear and the scaling factor is actually constant, as depicted in Figure~\ref{fig-framework}. Detailed discussion is provided in Appendix~\ref{sec-formulate-3dvar-as-vae3dvar}.

\subsection{General Formulation of VAE-Var}

Our VAE-Var framework for data assimilation comprises two phases: training and assimilation. In the training phase (detailed in Algorithm~\ref{alg-training}), we follow the traditional NMC (National Meteorological Center) method to generate historical error samples and then use them to train a VAE. We note it here that although in real-world scenarios, $\mathcal{M}^{gt}$ in Algorithm~\ref{alg-training} is generally not available, the reanalysis data can be utilized to approximate $\mathcal{M}_{0\to\tau}^{gt}(\mathbf{x}_0)$. Once the VAE is trained, we employ the decoder of the VAE for implementing the VAE-Var assimilation algorithm (outlined in Algorithm~\ref{alg-assimilation}).

\begin{minipage}{.43\textwidth}
\begin{algorithm}[H]
\centering
\caption{Training Set Construction}
\label{alg-training}
\begin{algorithmic}
\Require: Ground truth model $\mathcal{M}^{gt}$, prediction model $\mathcal{M}$, sample number $N$, time gap $\tau$

\For{$i$ from $1$ to $N$}
    \State Randomly generate $\mathbf{x}_0$
    \State $\mathbf{x}_1 \gets \mathcal{M}_{\tau\to2\tau}\circ\mathcal{M}_{0\to\tau}^{gt}(\mathbf{x}_0)$
    \State $\mathbf{x}_2 \gets \mathcal{M}_{\tau\to2\tau}\circ\mathcal{M}_{0\to\tau}(\mathbf{x}_0)$
    \State Add $\mathbf{x}_1 -\mathbf{x}_2$ to the training set
\EndFor

\end{algorithmic}
\end{algorithm}  
\end{minipage}
\hfill
\begin{minipage}{.53\textwidth}
\begin{algorithm}[H]
\caption{VAE-Var Assimilation}
\label{alg-assimilation}
\begin{algorithmic}
\Require: Trained decoder $\mathcal{D}$, background state $\mathbf{x}_b$, observation $\mathbf{y}$

\State $\mathbf{z} \gets \mathbf{0}$ 
\State $\tilde{\mathcal{L}}_b^{vae}(\mathbf{z}) \gets \frac12 \mathbf{z}^\mathrm{T} \mathbf{z} + \frac12 \log \det\left(\frac{\partial \mathcal{D}(\mathbf{z})}{\partial \mathbf{z}}\right)^\mathrm{T}\left(\frac{\partial \mathcal{D}(\mathbf{z})}{\partial \mathbf{z}}\right)$
\State $\tilde{\mathcal{L}}_o^{vae}(\mathbf{z}) \gets \mathcal{L}_o^{vae}\left(\mathcal{D}(\mathbf{z}) + \mathbf{x}_b, \mathbf{y}\right)$
\State Minimize $\tilde{\mathcal{L}}^{vae}(\mathbf{z}) = \tilde{\mathcal{L}}_b^{vae}(\mathbf{z}) + \tilde{\mathcal{L}}_o^{vae}(\mathbf{z})$ with L-BFGS~\citep{liu1989limited, paszke2017automatic} and get the minimum point $\mathbf{z}^\star$
\State $\mathbf{x}_a = \mathcal{D}(\mathbf{z}^\star) + \mathbf{x}_b$

\end{algorithmic}
\end{algorithm}  
\end{minipage}

It is important to highlight that our algorithm enhances the background term of variational assimilation methods, making it suitable not only for 3DVar but also for 4DVar. If we use Equation~\ref{eq-3dvar-obs} to calculate $\tilde{\mathcal{L}}_o(\mathbf{z})$, it corresponds to a 3D version of VAE-Var, which we refer to as \textbf{VAE-3DVar}. Similarly, if Equation~\ref{eq-4dvar-obs} is used instead, the resulting algorithm is named \textbf{VAE-4DVar}. The visualization of implementing VAE-3DVar is also demonstrated in the lower part of Figure~\ref{fig-framework}.

\subsection{Implementing VAE-Var on Low-Dimensional Systems}

While our proposed VAE-Var framework theoretically applies to any dynamical system, this paper primarily focuses on its implementation in low-dimensional systems.

\paragraph{Calculation of the Jacobian Determinant}

Computing the Jacobian determinant for an arbitrary neural network $\mathcal{D}$ is challenging. Given our focus on low-dimensional dynamical systems, multi-layer perceptrons (MLPs) are adequate for constructing VAEs to learn their background error distributions. Take the three-layer perceptron as an example. The neural network $\mathcal{D}$ can be expressed as
\begin{equation}
    \mathcal{D}(\mathbf{z}) = \mathbf{A}_3 \alpha_2\left( \mathbf{A}_2 \alpha_1\left( \mathbf{A}_1 \mathbf{z} + \mathbf{b}_1\right)+ \mathbf{b}_2 \right)+ \mathbf{b}_3,
\end{equation}
where $\mathbf{A}_1, \mathbf{A}_2, \mathbf{A}_3, \mathbf{b}_1, \mathbf{b}_2, \mathbf{b}_3$ correspond to the weights and biases of three linear layers and $\alpha_1(\cdot)$, $\alpha_2(\cdot)$ correspond to the activation functions. The Jacobian can then be explicitly calculated as follows: 
\begin{equation}
    \frac{\partial \mathcal{D}(\mathbf{z})}{\partial \mathbf{z}} = \mathbf{A}_3 \mathrm{diag} \left(\alpha_2^\prime \left( \mathbf{A}_2\alpha_1(\mathbf{A}_1\mathbf{z}+\mathbf{b}_1) + \mathbf{b}_2 \right)\right) \mathbf{A}_2 \mathrm{diag}\left(\alpha_1^\prime \left( \mathbf{A}_1\mathbf{z} + \mathbf{b}_1 \right)\right) \mathbf{A}_1.
\end{equation}
The determinant $\det\left(\frac{\partial \mathcal{D}(\mathbf{z})}{\partial \mathbf{z}}\right)^\mathrm{T}\left(\frac{\partial \mathcal{D}(\mathbf{z})}{\partial \mathbf{z}}\right)$ can then be calculated using the PyTorch package~\citep{paszke2019pytorch}. 

\paragraph{Stabilizing the Determinant} 

Due to the high non-linearity of the neural network $\mathcal{D}$, the determinant function $\det\left(\frac{\partial \mathcal{D}(\mathbf{z})}{\partial \mathbf{z}}\right)^\mathrm{T}\left(\frac{\partial \mathcal{D}(\mathbf{z})}{\partial \mathbf{z}}\right)$ can become unstable and difficult to optimize. To stabilize the optimization process, we incorporate a diagonal matrix as a regularization term into the Jacobian before computing the determinant. This results in the modified background term:
\begin{equation}
\label{eq-bg-cost-total}
    \tilde{\mathcal{L}}_b^{vae}(\mathbf{z}) = \frac12 \mathbf{z}^\mathrm{T}\mathbf{z} + \frac12 \log \det\left(\left(\frac{\partial \mathcal{D}(\mathbf{z})}{\partial \mathbf{z}}\right)^\mathrm{T}\left(\frac{\partial \mathcal{D}(\mathbf{z})}{\partial \mathbf{z}}\right) + \epsilon \mathbf{I} \right),
\end{equation}
where $\mathbf{I}$ is the identity matrix and $\epsilon$ is a small positive number.

\section{Experiments}

\subsection{Experimental Setup} 

\paragraph{Experimental Design} Our experiments are conducted on low-dimensional chaotic systems with simulated observations and background states. First, we generate $N$ random physical states from the standard normal distribution and divide them into two parts $N = N_{train} + N_{val}$. The first $N_{train}$ samples are used for generating the training set according to Algorithm~\ref{alg-training}. The last $N_{val}$ samples are reserved for evaluating the assimilation algorithm. 
For each of these $N_{val}$ samples, we integrate from it using the ground truth model for $\tau$ steps to do a warm-up. Then, we integrate using the prediction model and the ground truth model for $\tau$ (in Algorithm~\ref{alg-training}) steps to obtain the background state $\mathbf{x}_b$, and the ground truth state $\mathbf{x}_{gt}$, respectively. After that, in the 3DVar case, we simulate the observations by specifying a linear or nonlinear observation operator $\mathcal{H}$ and introducing Gaussian noise $\epsilon_{noise}$, that is $\mathbf{y} = \mathcal{H}(\mathbf{x}_{gt})+\epsilon_{noise}$. The 4DVar case is very similar and the difference is that we need to generate a time sequence of observations (detailed in Appendix~\ref{sec-4dvar-detail}). By applying Algorithm~\ref{alg-assimilation}, we perform a VAE-Var assimilation to calculate the analysis state $\mathbf{x}_{a}$ using the background state $\mathbf{x}_b$ and the observations $\mathbf{y}$. In all our experiments, $N_{train} = 4000$ and $N_{val} = 1000$.

\paragraph{Evaluation} We assess our VAE-Var framework by calculating the rooted-mean-squared error (RMSE) between the analysis state $\mathbf{x}_{a}$ and the ground truth state $\mathbf{x}_{gt}$ and averaging them over all $N_{val}$ samples. Our framework is compared with the traditional variational assimilation algorithm. We also design another metric, $\mathrm{Imp}$, to evaluate the accuracy gain of the proposed VAE-Var algorithm. Denote $R_{bg}$, $R_{vae}$, $R_{trad}$ the RMSE of (1) the background state, (2) the analysis state calculated by VAE-Var, and (3) the analysis state calculated by the traditional algorithm, respectively. Then, $\mathrm{Imp}$ is defined as $\mathrm{Imp} = \frac{R_{bg} - R_{vae}}{R_{bg} - R_{trad}} - 1$, where $R_{bg} - R_{trad}$ and $R_{bg} - R_{var}$ measure the assimilation gain of the traditional algorithm and our algorithm, respectively. If $\mathrm{Imp}$ is larger than zero, that is, $R_{bg} - R_{var} > R_{bg} - R_{trad}$, it means that VAE-Var outperforms the traditional algorithm. 

\paragraph{Dynamical Systems} Two dynamical systems are chosen: Lorenz 63~\citep{lorenz1963deterministic} and Lorenz 96~\citep{lorenz1996predictability}. These two systems have been widely used for the evaluation of various data assimilation algorithms, owing to their chaotic properties and similarities with meteorological systems, especially in data-driven and machine learning case studies~\citep{lguensat2017analog, raissi2018deep, fablet2021learning, frerix2021variational}. The Lorenz 63 and Lorenz 96 systems are governed by ordinary differential equations (ODEs) formulated in Equation~\ref{eq-Lorenz-63} and Equation~\ref{eq-Lorenz-96}, respectively.

\begin{minipage}{.45\textwidth}
\begin{equation}
    \label{eq-Lorenz-63}
    \left\{
    \begin{aligned}
    \frac{\dif X}{\dif t} &= \sigma (Y - X) \\
    \frac{\dif Y}{\dif t} &= X (\rho - Z) -Y \\
    \frac{\dif Z}{\dif t} &= XY - \beta Z \\
    \end{aligned}
    \right.
\end{equation}
\end{minipage}
\hfill
\begin{minipage}{.50\textwidth}
\begin{equation}
    \label{eq-Lorenz-96}
    \begin{aligned}
    \frac{\dif X_i}{\dif t} = \left(X_{i+1} - X_{i-2}\right)X_{i-1} - X_i + F \\
    (1\leq i \leq d, X_{-1} = X_{d-1}, X_{0} = X_{d}) \\
    \end{aligned}
\end{equation}
\end{minipage}

\subsection{3DVar Results}

\subsubsection{Lorenz 63 System}

Parameter values $\sigma = 10$, $\rho = 28$, and $\beta = \frac{8}{3}$ are used to define the ground truth model $\mathcal{M}_{gt}$. Integration is performed with a step size of 0.01 over $\tau = 10$ steps to generate the training set and background states. In this section, we perturb the parameter $\sigma$ from 10 to 11 to construct the prediction model (Please refer to Appendix~\ref{sec-additional-results} for parameter selection explanation.). A three-layer perceptron is used for constructing VAE's encoder and decoder. Parameters of network structures and training details are provided in Appendix~\ref{sec-exp-detail}. The positive constant $\epsilon$ in Equation~\ref{eq-bg-cost-total} is set to 0.01. 

\paragraph{Linear Observation Operator} In this experiment, we construct the observations by selectively observing certain variables and masking others, resulting in linear observation operators. For example, setting $\mathcal{H}(\mathbf{x}) = \left( \begin{smallmatrix} 1 & 0 & 0 \\ 0 & 1 & 0 \end{smallmatrix}\right) \mathbf{x}$ indicates that   variables "$X$" and "$Y$" are observed with an "identity" function. The variance of the Gaussian noise $\epsilon_{noise}$ is set to $\sigma_{noise}^2\mathbf{I}$, where $\sigma_{noise}$ ranges from 0.1 to 0.5 in steps of 0.01. 
We select three observation mask scenarios and present the experimental results in the first and the second rows of Figure~\ref{fig-L63-sigma}, with additional results for other observation mask cases included in Appendix~\ref{sec-additional-results}. Additionally, we conduct an ablation study by removing the scaling term, both the scaling term and the regularization term, and report the results in the same panel.
By comparing the RMSE of the traditional 3DVar (blue line) and VAE-3DVar (purple line), our proposed method consistently outperforms the traditional algorithm across all observational settings. Particularly, as observation noise increases, the accuracy improvement of our method becomes more pronounced. This underscores the importance of leveraging the non-Gaussian structure of background information for effective assimilation when observations are less accurate.
The ablation study (represented by the orange line, green line, and purple line) demonstrates that all three terms in the variational cost contribute to enhancing assimilation accuracy. Utilizing only the observation term (orange line) can result in notably poor performance, potentially even worse than the "Naive Substitute" method. Introducing the regularization term (green line) leads to a significant improvement in performance, and further enhancement is achieved by introducing the scaling term (purple line).

\begin{figure}[ht]
    \centering
\includegraphics[width=1.0\linewidth]{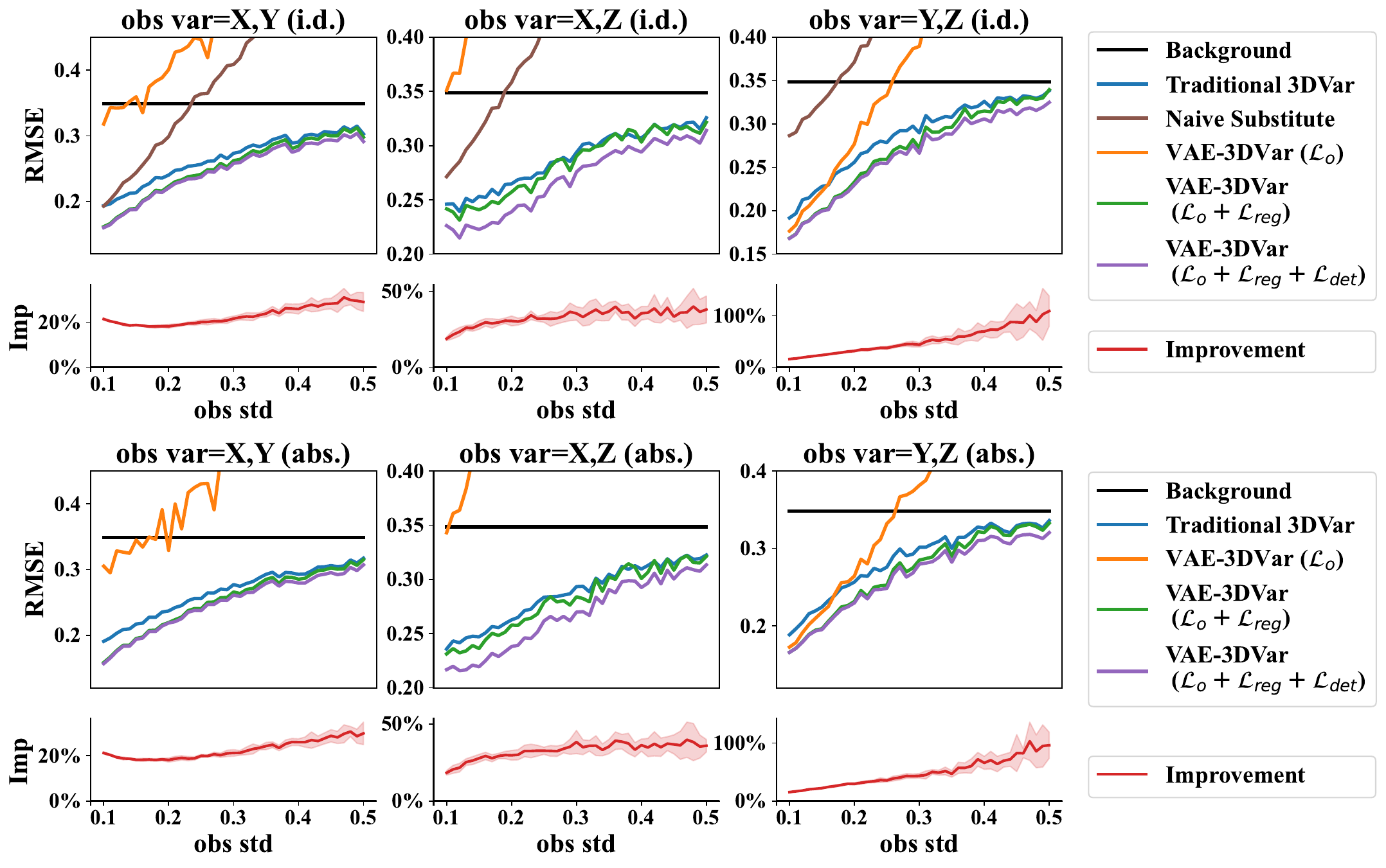}
    \caption{Results on the Lorenz 63 system with both linear (the first and the second rows) and nonlinear  (the third and the last rows) observation operators under 3DVar observational settings. Different panels correspond to different observational settings. For example, the title "obs var=X,Y (i.d.) / (abs.)" means that variables $X$ and $Y$ are observed with an "identity" or "absolute" function. The x-axis represents the standard deviation of the observation noise $\epsilon_{noise}$; the y-axis corresponds to two evaluation metrics: RMSE and $\mathrm{Imp}$. $\mathrm{Imp}$ is evaluated between the VAE-3DVar ($\mathcal{L}_o+\mathcal{L}_{reg}+\mathcal{L}_{det}$) and the traditional 3DVar. The experiments are repeated 10 times using different random noise, and we report the one-sigma error bars. The label "Naive Substitute" corresponds to the algorithm where the observed variables of the background state are replaced with their observed values. }
    \label{fig-L63-sigma}
\end{figure}

\paragraph{Non-linear Observation Operator} To evaluate the behaviour of VAE-Var under non-linear observation operators, we conduct experiments with an absolute function operator, that is, $\mathcal{H}(\mathbf{x}) = \vert \mathbf{H} \mathbf{x} \vert$. The experimental results are demonstrated in the third and the last rows of Figure~\ref{fig-L63-sigma}. Please refer to Appendix~\ref{sec-additional-results} for results with other observation masks. It can be found that the results under non-linear observational settings are consistent with our findings in the linear settings.


\subsubsection{Lorenz 96 System}

In Equation~\ref{eq-Lorenz-96}, the dimension $d$ of the Lorenz 96 system is set to 20 in our experiments. With regard to the ground truth model, $F$ is set to 8; as for the prediction model, we change $F$ in the first equation from 8 to 13 and keep other equations unchanged. A three-layer perceptron is also used for building the VAE, the details of which are provided in Appendix~\ref{sec-exp-detail}.

\begin{figure}[ht]
    \centering
\includegraphics[width=1.0\linewidth]{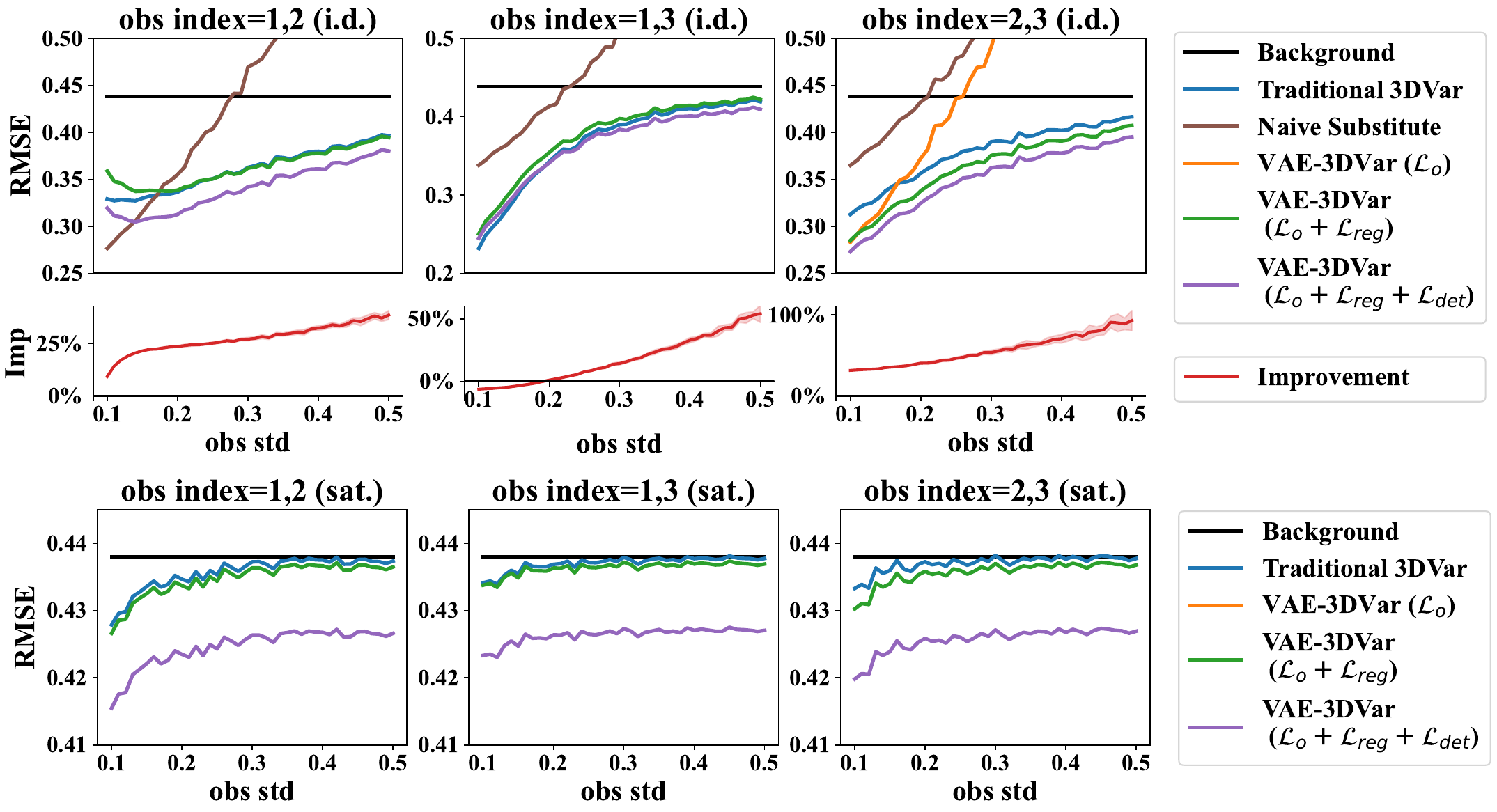}
    \caption{Results on the Lorenz 96 system with both linear (the first and the second rows) and nonlinear  (the last row) observation operators under 3DVar observational settings. In the title, "sat." is abbreviated for the saturated observation operator. The $\mathrm{Imp}$ metric is not demonstrated for nonlinear settings because $R_{bg} - R_{trad}$ can be very small, making the value of $\mathrm{Imp}$ larger than $10^4$. }
    \label{fig-L96-F}
\end{figure}

\paragraph{Linear Observation Operator} Similar to the experiments in the Lorenz 63 system, we construct the linear observation operator by observing only part of the variables. The variance of the Gaussian noise $\epsilon_{noise}$ is set to $\sigma_{noise}^2\mathbf{I}$, with $\sigma_{noise}$ ranging from 0.1 to 0.5 in steps of 0.01. We choose three different observation mask scenarios and present the experimental results in the first and the second rows of Figure~\ref{fig-L96-F}. The results demonstrate that VAE-3DVar generally outperforms the traditional 3DVar in most cases. In the rare instances where the observation noise is small and the traditional 3DVar performs better than VAE-Var, the difference is only marginal. Additionally, the incorporation of the regularization term and the scaling determinant term significantly contributes to the enhancement of the VAE-3DVar algorithm in the Lorenz 96 system.

\paragraph{Non-linear Observation Operator} In the Lorenz 96 system, we introduce a saturated function $f(x) = \frac{x}{1+|x|}$ to build the non-linear observation operator and demonstrate the results in Figure~\ref{fig-L96-F}. The comparison highlights a substantial improvement in accuracy with VAE-3DVar over the traditional 3DVar. This notable gain can be attributed to the nature of the saturated operator, which reduces the magnitude of observations and introduces increased noise, thereby emphasizing the critical need for more precise estimation of the background error.


\begin{figure}[ht]
    \centering
\includegraphics[width=1.0\linewidth]{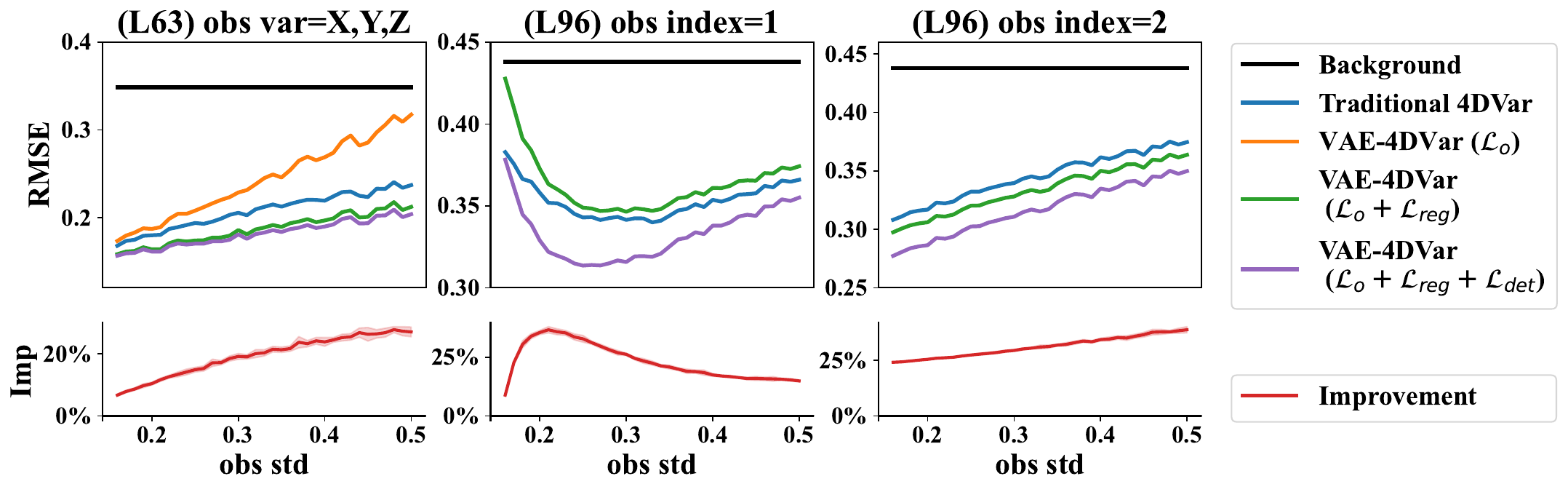}
    \caption{Results on the Lorenz 63 (the first column) and the Lorenz 96 (the second and the third columns) systems with linear observation operators under 4DVar observational setting. $\mathrm{Imp}$ is evaluated between VAE-4DVar ($\mathcal{L}_o+\mathcal{L}_{reg}+\mathcal{L}_{det}$) and traditional 4DVar. }
    \label{fig-4dvar-partial}
\end{figure}

\subsection{4DVar Results}

We have also conducted experiments under the 4DVar observational setting. The "assimilation window" length, corresponding to $n$ in Equation~\ref{eq-4dvar-obs}, is set to two and the interval between $t_i$ and $t_{i+1}$ is set to two integration steps. 
We utilize a linear observation operator and compare VAE-4DVar with traditional 4DVar, with the results depicted in Figure~\ref{fig-4dvar-partial}. This comparison also reveals an apparent improvement of VAE-4DVar over traditional 4DVar, with $\mathrm{Imp}$ ranging from 7\% to 40\%.

\section{Related Work}

The application of machine learning techniques to enhance traditional data assimilation algorithms has become a prominent and active research topic in recent years. Most of these works can be divided into two categories: latent-space assimilation and diffusion-based assimilation. 

\paragraph{Latent-space Assimilation} 
\cite{mack2020attention} is the first to propose using an autoencoder for dimension reduction and conducting assimilation in a low-dimensional latent space. They prove the equivalence of such an algorithm with the traditional 3DVar when the autoencoder is linear. This research primarily focuses on reducing the computational cost of traditional algorithms, with assimilation accuracy generally not showing improvement.
In \cite{amendola2020data} and \cite{melinc2023neural}, recurrent neural networks and variational autoencoders are employed to learn the latent space mapping. Although experimental results validate the effectiveness of these methods, they are not compared with traditional methods, and a rigorous mathematical formulation is also lacking.
In contrast, our work not only theoretically explains the superiority of our method over traditional methods but also validates our theory through experiments.

\paragraph{Diffusion-based Assimilation} In \cite{rozet2023score} and \cite{rozet2024score}, a diffusion model (SDA) is introduced for learning dynamical information, where observational data is incorporated using the diffusion posterior sampling method during inference. However, a drawback of this approach is that it does not accommodate the inclusion of background information. Following this work, \cite{huang2024diffda} presents the DiffDA method, which extends the diffusion model to learn non-Gaussian background error information. Despite this advancement, DiffDA is limited in its applicability to 4DVar and nonlinear observational settings. Moreover, both SDA and DiffDA are designed to sample from the distribution of the analysis state and cannot obtain the optimal estimate of the analysis state directly. By comparison, VAE-Var is specifically targeted at calculating the optimal state rather than sampling from its distribution.

\section{Conclusion}

In this paper, we address the limitation of traditional variational algorithms, which assume that background state errors follow a Gaussian distribution, and propose a novel data assimilation method named VAE-Var, which leverages VAE to learn the non-Gaussian characteristics of background error. We derive the rigorous expression for the variational cost and provide formulas for computing determinants in low-dimensional systems. Experimental results on two classical chaotic dynamical systems validate the effectiveness of our proposed method.

\paragraph{Limitation} The major limitation of our work relates to dimensionality. Learning from high-dimensional systems requires neural networks with a large number of parameters, which can render the computation of Jacobian determinants infeasible. Additionally, employing more complex neural network architectures beyond MLPs may pose challenges in obtaining closed-form Jacobian determinants, which we plan to address in future research.

\medskip

\bibliographystyle{unsrt}
\bibliography{ref}

\newpage

\appendix

\section{Broader Impact Statement}
\label{sec-broader-impact-state}

This paper presents our work aimed at extending the traditional data assimilation theorem with artificial intelligence to develop novel, more accurate assimilation algorithms. The development of a fundamental data assimilation theorem has significant societal implications across several key areas:

Improved data assimilation enhances the accuracy of climate predictions and weather forecasts, aiding in the prediction of extreme weather events and informing climate change mitigation strategies. This enables policymakers to make informed decisions to address global environmental challenges effectively.

Data assimilation also enhances agriculture by enabling precise monitoring of crop growth status and accurate yield estimation. By integrating data from satellite imagery, soil sensors, and weather forecasts, farmers can make real-time adjustments and informed decisions. This leads to optimal growth conditions, improved productivity, and better resource allocation.

In public health, advanced data assimilation in epidemiological models improves the prediction and management of disease outbreaks. During pandemics, reliable models inform public health responses and optimize resource allocation, ultimately saving lives by enabling timely interventions.

Furthermore, accurate economic models and forecasts derived from advanced data assimilation enhance risk management, inform fiscal policies, and support stable economic systems. This benefits businesses, governments, and individuals by promoting economic stability and growth.

Overall, the potential social benefits of this work extend to various sectors, and continued investment in this research is crucial for maximizing its broad impact.

\section{Derivation of the Background Term}
\label{sec-bg-term-derivation}

As mentioned in the main paper of this manuscript, we aim to simplify the following expression:
\begin{equation}
\begin{aligned}
    & \mathcal{L}_b^{vae}(\mathbf{x}, \mathbf{x}_b) \\
    = & -\lim_{\sigma_0\to 0}\log \int_{\mathbf{z}} \sigma_0^{-n} \exp{\left(-(\mathbf{x}-\mathbf{x}_b-\mathcal{D}(\mathbf{z}))^\mathrm{T}(\mathbf{x}-\mathbf{x}_b-\mathcal{D}(\mathbf{z}))/(2\sigma_0^2)\right)} \exp{\left(-\mathbf{z}^\mathrm{T}\mathbf{z}/2\right)} \dif \mathbf{z}.
\end{aligned}
\end{equation}

First, we do a variable transformation on the integral variable $\mathbf{z}$. Since $\mathcal{D}$ is bijective, let $\mathbf{z}=\mathcal{D}^{-1}(\mathbf{x}-\mathbf{x}_b+\sigma_0\mathbf{y})$, and then $\mathbf{y}$ becomes the new integral variable. Since we are calculating a volume integral, it is necessary to transform the differential element of volume from $\dif \mathbf{z}$ to $\dif \mathbf{y}$. In Chapter 2.5.3 of \cite{giaquinta2012mathematical}, the conversion formula of volume elements on manifolds embedded in Euclidean space of two different dimensions is given as
\begin{equation}
    \dif \mathbf{z} = \sigma_0^n \det\left(\left.\frac{\partial \mathcal{D}(\mathbf{z})}{\partial \mathbf{z}} \right|_{\mathbf{z}=\mathcal{D}^{-1}(\mathbf{x}-\mathbf{x}_b+\sigma_0\mathbf{y})}^\mathrm{T}  \left.\frac{\partial \mathcal{D}(\mathbf{z})}{\partial \mathbf{z}} \right|_{\mathbf{z}=\mathcal{D}^{-1}(\mathbf{x}-\mathbf{x}_b+\sigma_0\mathbf{y})} \right)^{-1/2} \dif \mathbf{y}.
\end{equation}
For simplicity, denote $\det^\star(x) = \sqrt{\det{x^\mathrm{T}x}}$. Then,
\begin{equation}
    \dif \mathbf{z} = \sigma_0^n\left( \left.\det\right.^\star \left.\frac{\partial \mathcal{D}(\mathbf{z})}{\partial \mathbf{z}} \right|_{\mathbf{z}=\mathcal{D}^{-1}(\mathbf{x}-\mathbf{x}_b+\sigma_0\mathbf{y})} \right)^{-1} \dif \mathbf{y}.
\end{equation}

We have
\begin{equation}
\begin{aligned}
    &\mathcal{L}_b^{vae}(\mathbf{x}, \mathbf{x}_b) = -\lim_{\sigma_0\to 0} \log \int_\mathbf{z} \exp{\left(-(\mathbf{x}-\mathbf{x}_b-\mathcal{D}(\mathbf{z}))^\mathrm{T}(\mathbf{x}-\mathbf{x}_b-\mathcal{D}(\mathbf{z}))/(2\sigma_0^2)\right)} \exp{\left(-\mathbf{z}^\mathrm{T}\mathbf{z}/2\right)} \dif \mathbf{z} \\
    =& -\lim_{\sigma_0\to 0} \log \int_\mathbf{y} \exp{\left(-||\mathbf{y}||_2^2/2 - ||\mathcal{D}^{-1}(\mathbf{x}-\mathbf{x}_b+\sigma_0\mathbf{y})||_2^2/2\right)} \left( \left. \det\right.^\star \left.\frac{\partial \mathcal{D}(\mathbf{z})}{\partial \mathbf{z}} \right|_{\mathbf{z}=\mathcal{D}^{-1}(\mathbf{x}-\mathbf{x}_b+\sigma_0\mathbf{y})} \right)^{-1} \dif \mathbf{y}.
\end{aligned}
\end{equation}

Denote
\begin{equation}
    \phi(\mathbf{y}; \sigma_0) = \exp{\left(-||\mathbf{y}||_2^2/2 - ||\mathcal{D}^{-1}(\mathbf{x}-\mathbf{x}_b+\sigma_0\mathbf{y})||_2^2/2\right)} \left( \left. \det\right.^\star \left.\frac{\partial \mathcal{D}(\mathbf{z})}{\partial \mathbf{z}} \right|_{\mathbf{z}=\mathcal{D}^{-1}(\mathbf{x}-\mathbf{x}_b+\sigma_0\mathbf{y})} \right)^{-1}
\end{equation}
and it is obvious that
\begin{equation}
\begin{aligned}
    \lim_{\sigma_0\to 0} \phi(\mathbf{y}; \sigma_0) = \phi(\mathbf{y}; 0) = \exp{\left(-||\mathbf{y}||_2^2/2 - ||\mathcal{D}^{-1}(\mathbf{x}-\mathbf{x}_b)||_2^2/2\right)} \left( \left. \det\right.^\star \left.\frac{\partial \mathcal{D}(\mathbf{z})}{\partial \mathbf{z}} \right|_{\mathbf{z}=\mathcal{D}^{-1}(\mathbf{x}-\mathbf{x}_b)} \right)^{-1}.
\end{aligned}
\end{equation}

From the assumption that $\det\left(\left(\frac{\partial \mathcal{D}(\mathbf{z})}{\partial \mathbf{z}}\right)^\mathrm{T}\left(\frac{\partial \mathcal{D}(\mathbf{z})}{\partial \mathbf{z}}\right)\right) > C$, we have
\begin{equation}
    \left( \left. \det\right.^\star \left.\frac{\partial \mathcal{D}(\mathbf{z})}{\partial \mathbf{z}} \right|_{\mathbf{z}=\mathcal{D}^{-1}(\mathbf{x}-\mathbf{x}_b+\sigma_0\mathbf{y})} \right)^{-1} < C^{-1/2}.
\end{equation}
Also, it is apparent that 
\begin{equation}
    \exp{\left(-||\mathbf{y}||_2^2/2 - ||\mathcal{D}^{-1}(\mathbf{x}-\mathbf{x}_b+\sigma_0\mathbf{y})||_2^2/2\right)} < 1.
\end{equation}
Hence, $\phi(\mathbf{y}; \sigma_0) < C^{-1/2}$; that is, $\phi(\mathbf{y}; \sigma_0)$ is consistently bounded. According to the Lebesgue's dominated convergence theorem, the order of integral and limit can be exchanged. Therefore, we have:
\begin{equation}
\begin{aligned}
& \lim_{\sigma_0\to 0} \int_\mathbf{y} \phi(\mathbf{y}; \sigma_0) \dif \mathbf{y} = \int_\mathbf{y} \lim_{\sigma_0\to 0} \phi(\mathbf{y}; \sigma_0) \dif \mathbf{y} \\
= & \int_\mathbf{y} \exp{\left(-||\mathbf{y}||_2^2/2 - ||\mathcal{D}^{-1}(\mathbf{x}-\mathbf{x}_b)||_2^2/2\right)} \left( \left. \det\right.^\star \left.\frac{\partial \mathcal{D}(\mathbf{z})}{\partial \mathbf{z}} \right|_{\mathbf{z}=\mathcal{D}^{-1}(\mathbf{x}-\mathbf{x}_b)} \right)^{-1} \dif \mathbf{y} \\
=& \exp{\left(- ||\mathcal{D}^{-1}(\mathbf{x}-\mathbf{x}_b)||_2^2/2\right)} \left( \left. \det\right.^\star \left.\frac{\partial \mathcal{D}(\mathbf{z})}{\partial \mathbf{z}} \right|_{\mathbf{z}=\mathcal{D}^{-1}(\mathbf{x}-\mathbf{x}_b)} \right)^{-1} + Constant. \\
\end{aligned}
\end{equation}
This implies that the limit $\lim_{\sigma_0\to 0} \int_\mathbf{y} \phi(\mathbf{y}; \sigma_0) \dif\mathbf{y}$ exists and is larger than zero. Therefore the order of logarithm and limit can be exchanged. That is,
\begin{equation}
\begin{aligned}
\mathcal{L}_b^{vae}(\mathbf{x}, \mathbf{x}_b) &= -\lim_{\sigma_0\to 0} \log \int_\mathbf{y} \phi(\mathbf{y}; \sigma_0) \dif\mathbf{y} \\
&= -\log \lim_{\sigma_0\to 0} \int_\mathbf{y} \phi(\mathbf{y}; \sigma_0) \dif\mathbf{y} \\
&= -\log \exp{\left(- ||\mathcal{D}^{-1}(\mathbf{x}-\mathbf{x}_b)||_2^2/2\right)} \left( \left. \det\right.^\star \left.\frac{\partial \mathcal{D}(\mathbf{z})}{\partial \mathbf{z}} \right|_{\mathbf{z}=\mathcal{D}^{-1}(\mathbf{x}-\mathbf{x}_b)} \right)^{-1} \\
&= \frac12 ||\mathcal{D}^{-1}(\mathbf{x}-\mathbf{x}_b)||_2^2 + \log \left. \det\right.^\star \left.\frac{\partial \mathcal{D}(\mathbf{z})}{\partial \mathbf{z}} \right|_{\mathbf{z}=\mathcal{D}^{-1}(\mathbf{x}-\mathbf{x}_b)}. \\
\end{aligned}
\end{equation}

Let $\mathbf{z}=\mathcal{D}^{-1}(\mathbf{x}-\mathbf{x}_b)$ and then 
\begin{equation}
\begin{aligned}
\tilde{\mathcal{L}}_b^{vae}(\mathbf{z}) = & \frac12 ||\mathcal{D}^{-1}(\mathbf{x}-\mathbf{x}_b)||_2^2 + \log \left. \det\right.^\star \left.\frac{\partial \mathcal{D}(\mathbf{z})}{\partial \mathbf{z}} \right|_{\mathbf{z}=\mathcal{D}^{-1}(\mathbf{x}-\mathbf{x}_b)} \\
= & \frac12 ||\mathbf{z}||_2^2 + \frac12 \log \det \left( \frac{\partial \mathcal{D}(\mathbf{z})}{\partial \mathbf{z}} \right)^\mathrm{T} \left( \frac{\partial \mathcal{D}(\mathbf{z})}{\partial \mathbf{z}} \right), \\
\end{aligned}
\end{equation}
which concludes the proof.

\section{Details of 4DVar}

\label{sec-4dvar-detail}

According to the Bayes' Theorem,
\begin{equation}
    \arg \max_{\mathbf{x}} p(\mathbf{x}|\mathbf{y}_0, ...\mathbf{y}_{N-1}) = \arg \max_{\mathbf{x}} \frac{p(\mathbf{y}_0, ...\mathbf{y}_{N-1}|\mathbf{x}) p(\mathbf{x})}{p(\mathbf{y}_0, ...\mathbf{y}_{N-1})} = \arg \max_{\mathbf{x}} p(\mathbf{y}_0, ...\mathbf{y}_{N-1}|\mathbf{x}) p(\mathbf{x}).
\end{equation}
We define the variational cost as:
\begin{equation}
    \mathcal{L}(\mathbf{x}) = -\log p(\mathbf{y}_0, ...\mathbf{y}_{N-1}|\mathbf{x}) p(\mathbf{x}) = - \log p(\mathbf{y}_0, ...\mathbf{y}_{N-1}|\mathbf{x}) - \log p(\mathbf{x}).
\end{equation}
The observation term is defined as $\mathcal{L}_o(\mathbf{x}, \mathbf{y}_0, ...\mathbf{y}_{N-1}) = - \log p(\mathbf{y}_0, ...\mathbf{y}_{N-1}|\mathbf{x})$ and the background term is defined as $\mathcal{L}_b(\mathbf{x}, \mathbf{x}_b) = - \log p(\mathbf{x})$. The background term is identical to that in 3DVar and we only need to address the observation term. It is assumed that the observations at different time steps are independent. Therefore, the observation term can be decomposed as:
\begin{equation}
    \mathcal{L}_o(\mathbf{x}, \mathbf{y}_0, ...\mathbf{y}_{N-1}) = -\log p(\mathbf{y}_0, ...\mathbf{y}_{N-1}|\mathbf{x}) = - \sum_{n=0}^{N-1}\log p(\mathbf{y}_n|\mathbf{x}).
\end{equation}
Supposing that the prediction model $\mathcal{M}$ has no error, the conditional distribution $\mathbf{y}_n|\mathbf{x}$ follows a Gaussian distribution $\mathcal{N}\left(\mathcal{H}\left(\mathcal{M}_{0\to n}(\mathbf{x})\right), \mathbf{R}\right)$. Therefore, we have
\begin{equation}
    \mathcal{L}_o(\mathbf{x}, \mathbf{y}_0, ..., \mathbf{y}_{N-1}) = \frac12 \sum_{n=0}^{N-1} (\mathbf{y} - \mathcal{H}(\mathcal{M}_{0\to n}(\mathbf{x})))^\mathrm{T} \mathbf{R}^{-1} (\mathbf{y} - \mathcal{H}(\mathcal{M}_{0\to n}(\mathbf{x}))).
\end{equation}

\paragraph{Simulated Observation Generation} 
In our simulated experiments, we begin by generating a ground truth state, \(\mathbf{x}_{gt(0)} = \mathbf{x}_{gt}\). We then integrate this state using the ground truth model \(\mathcal{M}_{gt}\) to obtain the ground truth states at future time steps, \(\mathbf{x}_{gt(n)} = \mathcal{M}_{gt(0\to n)}(\mathbf{x}_{gt})\). Next, we specify an observation operator \(\mathcal{H}\) and introduce Gaussian noise \(\epsilon_{noise}\) to construct the observation series \(\mathbf{y}_n = \mathcal{H}(\mathbf{x}_{gt(n)}) + \epsilon_{noise}\).

\section{Experimental Details}

\label{sec-exp-detail}

\paragraph{Resources} All experiments were conducted on a MacBook Pro laptop using CPUs.

\begin{figure}[ht]
    \centering
\includegraphics[width=1.0\linewidth]{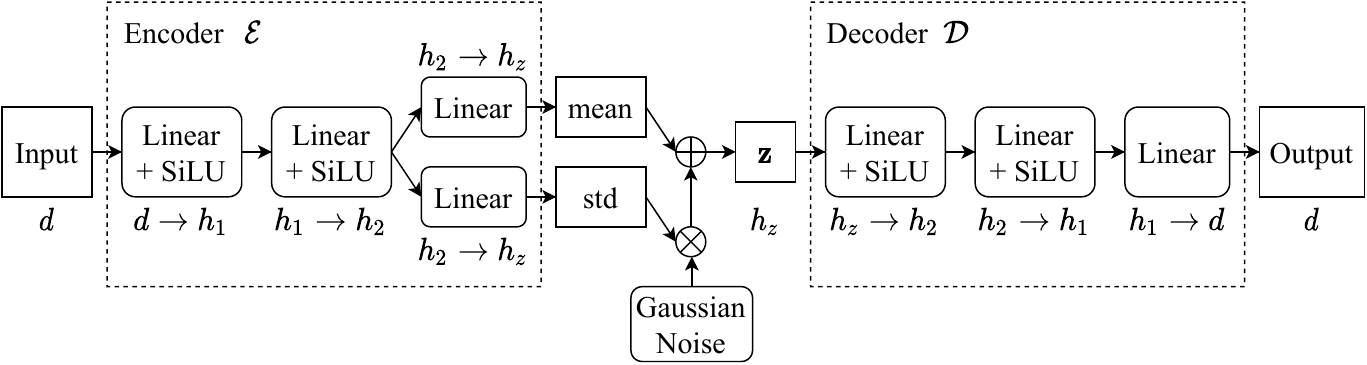}
    \caption{Structure of the variational autoencoder. }
    \label{fig-vae-structure}
\end{figure}

\paragraph{Network Details} For both the Lorenz 63 and Lorenz 96 systems, the network structures for constructing the encoder and the decoder are three-layer perceptrons, utilizing SiLU~\citep{elfwing2018sigmoid} as the activation function, as illustrated in Figure~\ref{fig-vae-structure}. We train the neural network with the AdamW optimizer~\citep{loshchilov2017decoupled}. Different hyperparameters are used for learning the background errors of different ODEs with perturbed parameters. Detailed information on these hyperparameters is provided in Table~\ref{tab-hyperparameters}.

\begin{table}
\centering
\caption{Parameters of the VAE structure and training.  } 
\centering
\begin{tabular}{cccc}
\toprule
    Dynamical system & \multicolumn{2}{c}{Lorenz 63} & Lorenz 96 \\
    \cmidrule(r){1-1} \cmidrule(r){2-3} \cmidrule(r){4-4} 
    Parameters perturbed & $\sigma$ & $\rho$ & $F$ \\
\midrule
    $h_1$      &   8 &  10 &  35 \\ 
    $h_2$      &   8 &  10 &  35 \\ 
    $h_z$      &   3 &   5 &  15 \\ 
    $\sigma_0$ & 0.3 & 0.2 & 0.1 \\ 
Learning rate  & $10^{-3}$ & $10^{-3}$ & $10^{-3}$ \\ 
Epochs         & 300 & 300 & 1000 \\
Batch size     &  32 &  32 &  32  \\
$\epsilon$     & $10^{-2}$ & $10^{-5}$ & $10^{-2}$ \\
\bottomrule
\end{tabular}
\label{tab-hyperparameters}
\end{table}

\section{Formulate 3DVar as a Special Case of VAE-3DVar}

\label{sec-formulate-3dvar-as-vae3dvar}

In this section, we aim to demonstrate that when a full-rank linear function is used to construct the decoder \(\mathcal{D}\) of VAE-3DVar, the formulation of our assimilation process is equivalent to the traditional 3DVar.

Assume \(\mathcal{D}(\mathbf{x}) = \mathbf{A}\mathbf{x}\), where \(\mathbf{A}\) is a full-rank matrix. The background term in Equation~\ref{eq-vae-loss} can be simplified as follows:
\begin{equation}
\begin{aligned}
    \tilde{\mathcal{L}}_b^{vae}(\mathbf{z}) &= \frac12 \mathbf{z}^\mathrm{T}\mathbf{z} + \frac12 \log \det\left(\left(\frac{\partial \mathcal{D}(\mathbf{z})}{\partial \mathbf{z}}\right)^\mathrm{T}\left(\frac{\partial \mathcal{D}(\mathbf{z})}{\partial \mathbf{z}}\right)\right) \\
    &= \frac12 \mathbf{z}^\mathrm{T}\mathbf{z} + \frac12 \log \det\left(\mathbf{A}^\mathrm{T}\mathbf{A}\right) 
    = \frac12 \mathbf{z}^\mathrm{T}\mathbf{z} + \log \det\mathbf{A}.
\end{aligned}
\end{equation}
Since \(\mathbf{A}\) is a constant, it can be ignored during the optimization of the variational cost. Thus,
\begin{equation}
    \arg \min_{\mathbf{z}} \tilde{\mathcal{L}}_b^{vae}(\mathbf{z}) 
    = \arg \min_{\mathbf{z}} \left(\frac12 \mathbf{z}^\mathrm{T}\mathbf{z} + \log \det\mathbf{A}\right)
    = \arg \min_{\mathbf{z}} \frac12 \mathbf{z}^\mathrm{T}\mathbf{z}.
\end{equation}
This background term is identical to that of the traditional 3DVar. Since the observation terms are also the same, this suffices to prove that the formulation of VAE-Var is equivalent to the traditional variational algorithm when the decoder is a full-rank linear function.

It is important to note that the proof above only shows that the formulations are identical; it does not necessarily mean that the algorithms are identical. The key difference is that in our algorithm, we build a variational autoencoder and use the back-propagation algorithm to learn the parameters of \(\mathcal{D}\). In contrast, the traditional algorithm relies on expert knowledge~\citep{fisher2003background} about the matrix structure for the calculation of \(\mathbf{A}\).

\newpage

\section{Additional Results}

\label{sec-additional-results}

\subsection{Lorenz 63 (3DVar Settings)}

\subsubsection{Linear Observation Operator}

In the ODE of the Lorenz 63 system, there are three parameters: \(\sigma\), \(\rho\), and \(\beta\). Here, \(\sigma\) corresponds to the Rayleigh number, \(\rho\) corresponds to the Prandtl number, while \(\beta\) does not have a clear physical meaning. In our experiments, we perturb \(\sigma\) and \(\rho\) to generate different background state distributions. There are a total of seven possible observation mask scenarios, and we iterate through all of them to conduct the experiments.

\begin{figure}[H]
    \centering
\includegraphics[width=0.8\linewidth]{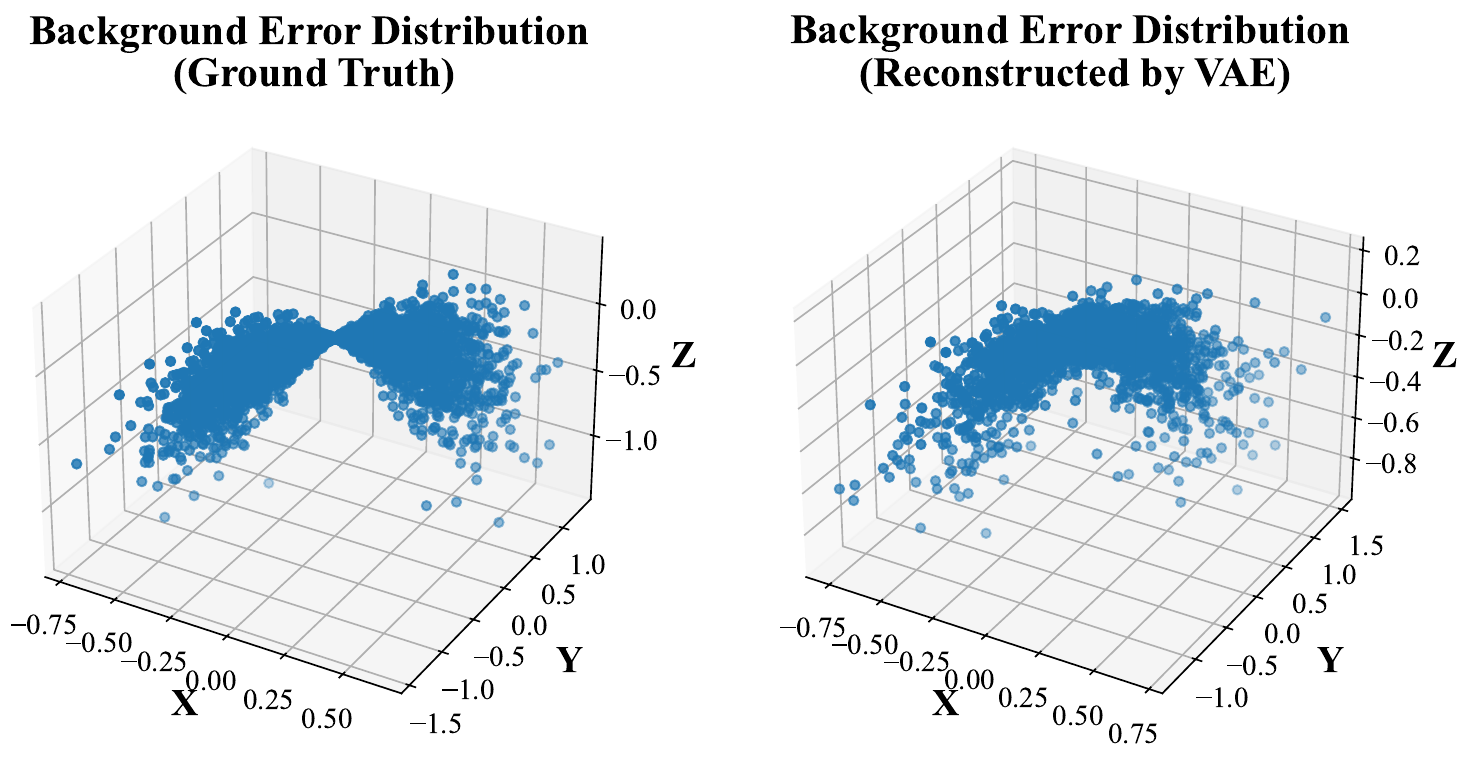}
    \caption{Background error distribution for the Lorenz 63 system when the ODE parameter is changed from $\sigma=10, \rho=28, \beta=\frac83$ to $\sigma=11, \rho=28, \beta=\frac83$. The left panel shows the ground truth background error distribution, and the right panel shows the background error distribution reconstructed by our trained VAE decoder. }
    \label{fig-bg-lorenz63-sigma}
\end{figure}

\begin{figure}[H]
    \centering
\includegraphics[width=0.8\linewidth]{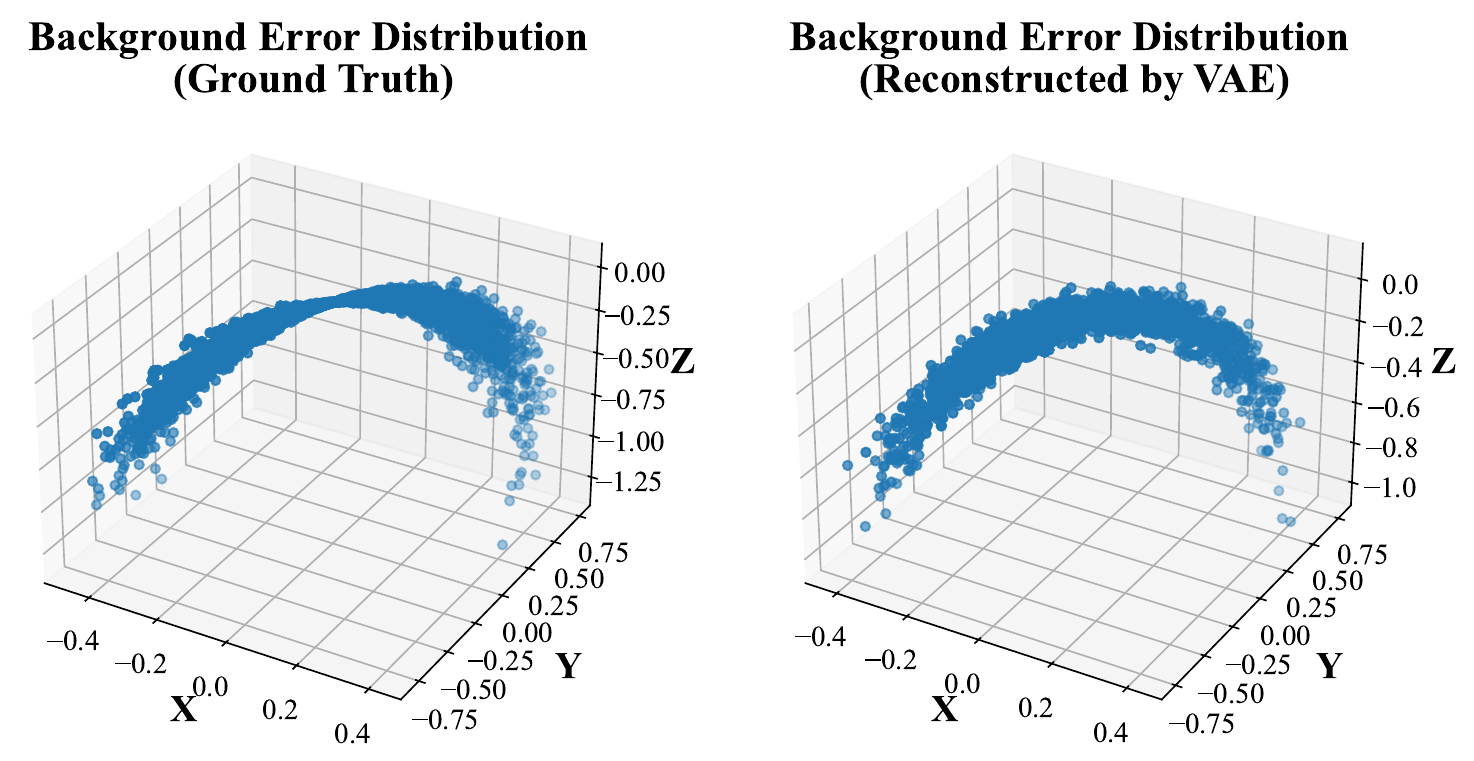}
    \caption{Background error distribution for the Lorenz 63 system when the ODE parameter is changed from $\sigma=10, \rho=28, \beta=\frac83$ to $\sigma=10, \rho=29, \beta=\frac83$. The left panel shows the ground truth background error distribution, and the right panel shows the background error distribution reconstructed by our trained VAE decoder. }
    \label{fig-bg-lorenz63-rho}
\end{figure}

\begin{figure}[H]
    \centering
\includegraphics[width=1.0\linewidth]{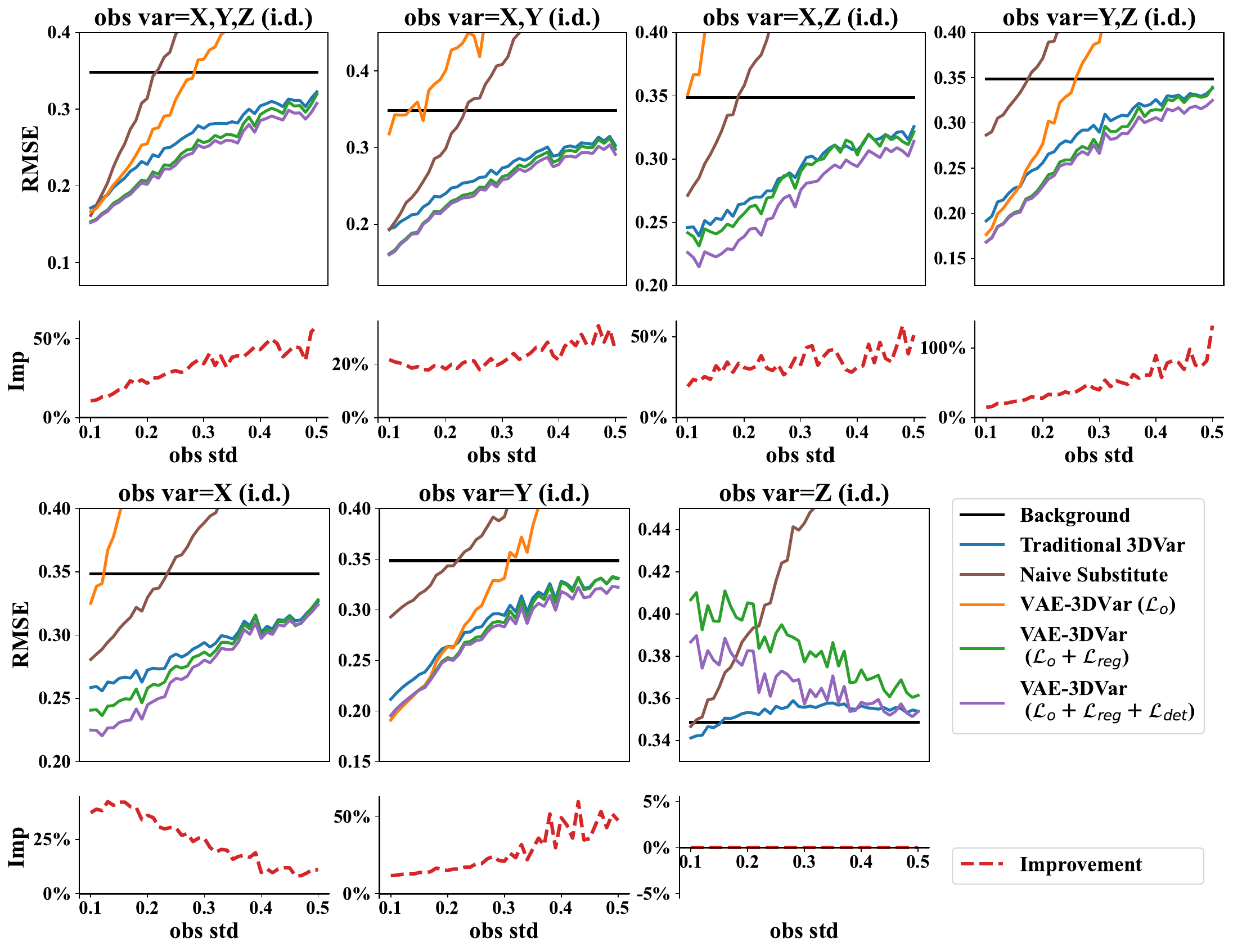}
    \caption{Results on the Lorenz 63 system with linear observation operators under 3DVar observational settings. The background states are constructed by changing the ODE parameter from $\sigma=10, \rho=28, \beta=\frac83$ to $\sigma=11, \rho=28, \beta=\frac83$. The metric $\mathrm{Imp}$ is set to zero if the analysis state after assimilation is worse than the background state.}
    \label{fig-lorenz63-3dvar-linear-appendix}
\end{figure}

\paragraph{Perturbing $\sigma$} In this experiment, we set $\sigma=11, \rho=28, \beta=\frac83$ to define the prediction model. Figure~\ref{fig-bg-lorenz63-sigma} demonstrates the background error distribution of perturbing $\sigma$, and Figure~\ref{fig-lorenz63-3dvar-linear-appendix} shows the comparison between different assimilation methods. 
It's evident that our proposed VAE-3DVar method outperforms the traditional 3DVar algorithm when the observed variables include "$X,Y,Z$", "$X,Y$", "$X,Z$", "$Y,Z$", "$X$", and "$Y$". In several of these scenarios, our improvement can reach as high as 100\%. However, our method does not perform well when only the variable $Z$ is observed. This is because when only the variable $Z$ is observed, the possible increment of $Y$ and $Z$ does not lie in a contiguous region. In Figure~\ref{fig-bg-lorenz63-sigma}, if we were to draw a plane parallel to the $X-Y$ plane, the intersection of this plane with the area represented by the blue dots intuitively splits into two separate parts: one on the left and the other on the right. Consequently, it becomes challenging for the assimilation algorithm to determine whether the increment should be allocated to the left or the right part. This inherent ambiguity, stemming from incomplete information, presents a difficulty for all the assimilation methods to achieve successful assimilation and, therefore, our method does not yield improvements in this particular case.

\begin{figure}[H]
    \centering
\includegraphics[width=1.0\linewidth]{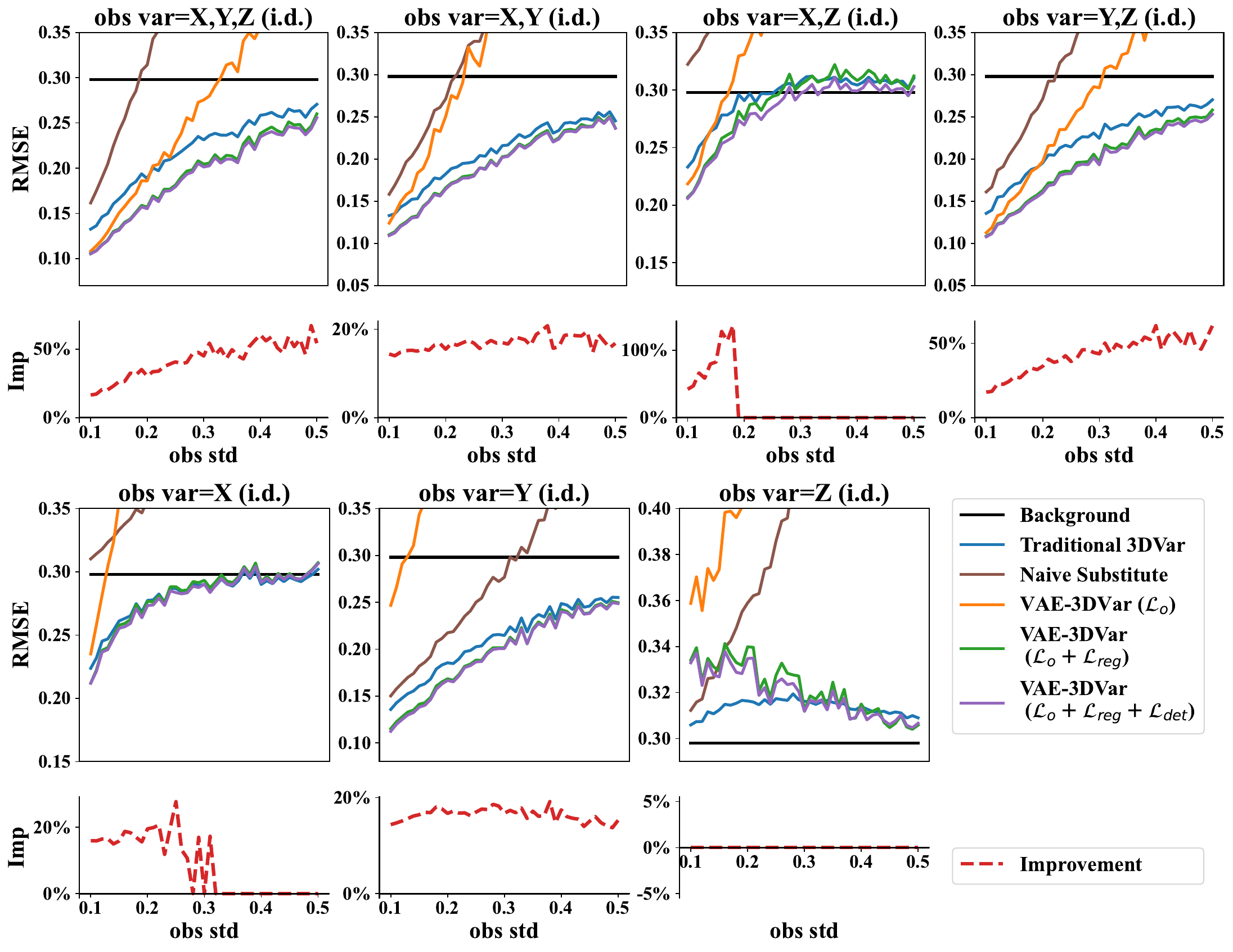}
    \caption{Results on the Lorenz 63 system with linear observation operators under 3DVar observational settings. The background states are constructed by changing the ODE parameter from $\sigma=10, \rho=28, \beta=\frac83$ to $\sigma=10, \rho=29, \beta=\frac83$. The metric $\mathrm{Imp}$ is set to zero if the analysis state after assimilation is worse than the background state. When the variable "$X$" and the variables "$X, Z$" are observed, $\mathrm{Imp}$ becomes zero if the observation noise is strong. This indicates that in these two cases, too strong observation noise will make the traditional algorithm ineffective. }
    \label{fig-lorenz63-3dvar-rho-appendix}
\end{figure}

\paragraph{Perturbing $\rho$} In this experiment, we set $\sigma=10, \rho=29, \beta=\frac83$ to define the prediction model. Figure~\ref{fig-bg-lorenz63-rho} demonstrates the background error distribution of perturbing $\sigma$, and Figure~\ref{fig-lorenz63-3dvar-rho-appendix} shows the comparison between different assimilation methods. The results are consistent with that of perturbing $\sigma$, with our method outperforming the traditional one when the observed variables are "$X,Y,Z$", "$X,Y$", "$X,Z$", "$Y,Z$", "$X$", or "$Y$". Similarly, when only $Z$ is observed, the issue of incomplete observational information accounts for the unusual results.

\subsubsection{Non-linear Observation Operator}

\begin{figure}[H]
    \centering
\includegraphics[width=1.0\linewidth]{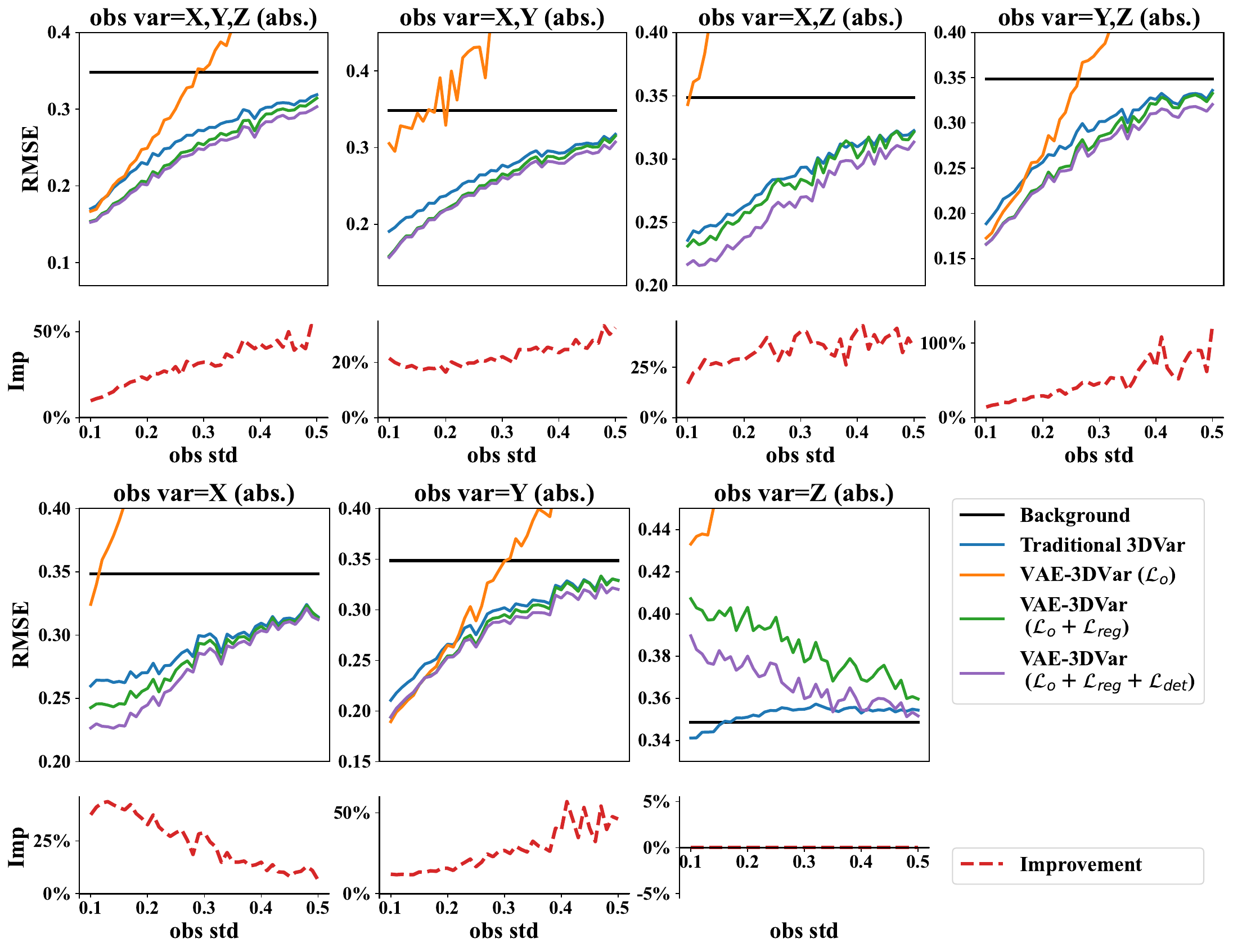}
    \caption{Results on the Lorenz 63 system with nonlinear observation operators under 3DVar observational settings. The observation operator is the absolute function. The background states are constructed by changing the ODE parameter from $\sigma=10, \rho=28, \beta=\frac83$ to $\sigma=11, \rho=28, \beta=\frac83$. The metric $\mathrm{Imp}$ is set to zero if the analysis state after assimilation is worse than the background state. }
    \label{fig-lorenz63-3dvar-nonlinear-appendix}
\end{figure}

In this experiment, we use the absolute function to construct the observation operator. The results for all seven different observation masks are demonstrated in Figure~\ref{fig-lorenz63-3dvar-nonlinear-appendix}. These results are consistent with those obtained using the linear observation operator.

\subsection{Lorenz 63 (4DVar Settings)}

\begin{figure}[H]
    \centering
\includegraphics[width=1.0\linewidth]{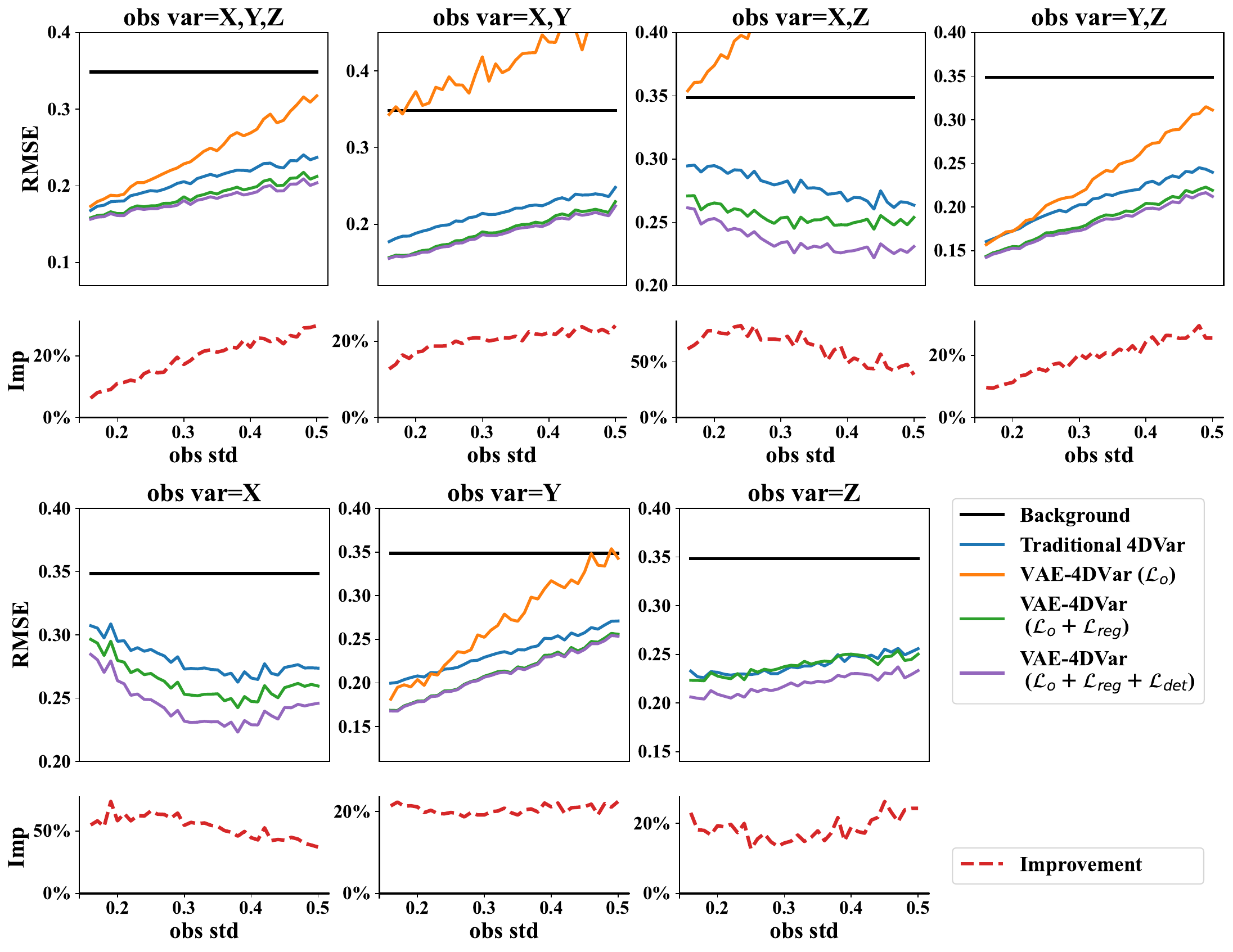}
    \caption{Results on the Lorenz 63 system with linear observation operators under 4DVar observational settings. The background states are constructed by changing the ODE parameter from $\sigma=10, \rho=28, \beta=\frac83$ to $\sigma=11, \rho=28, \beta=\frac83$. }
    \label{fig-lorenz63-4dvar-appendix}
\end{figure}

The results under the 4DVar observational settings with linear operators are demonstrated in Figure~\ref{fig-lorenz63-4dvar-appendix}. In this experiment, we set \(\sigma=11\), \(\rho=28\), and \(\beta=\frac{8}{3}\) to define the prediction model. It can be found that under all seven observation masks, our models outperform the traditional one, with improvements ranging from 7\% to over 70\%.

\subsection{Lorenz 96 (3DVar Settings)}

\begin{figure}[ht]
    \centering
\includegraphics[width=0.8\linewidth]{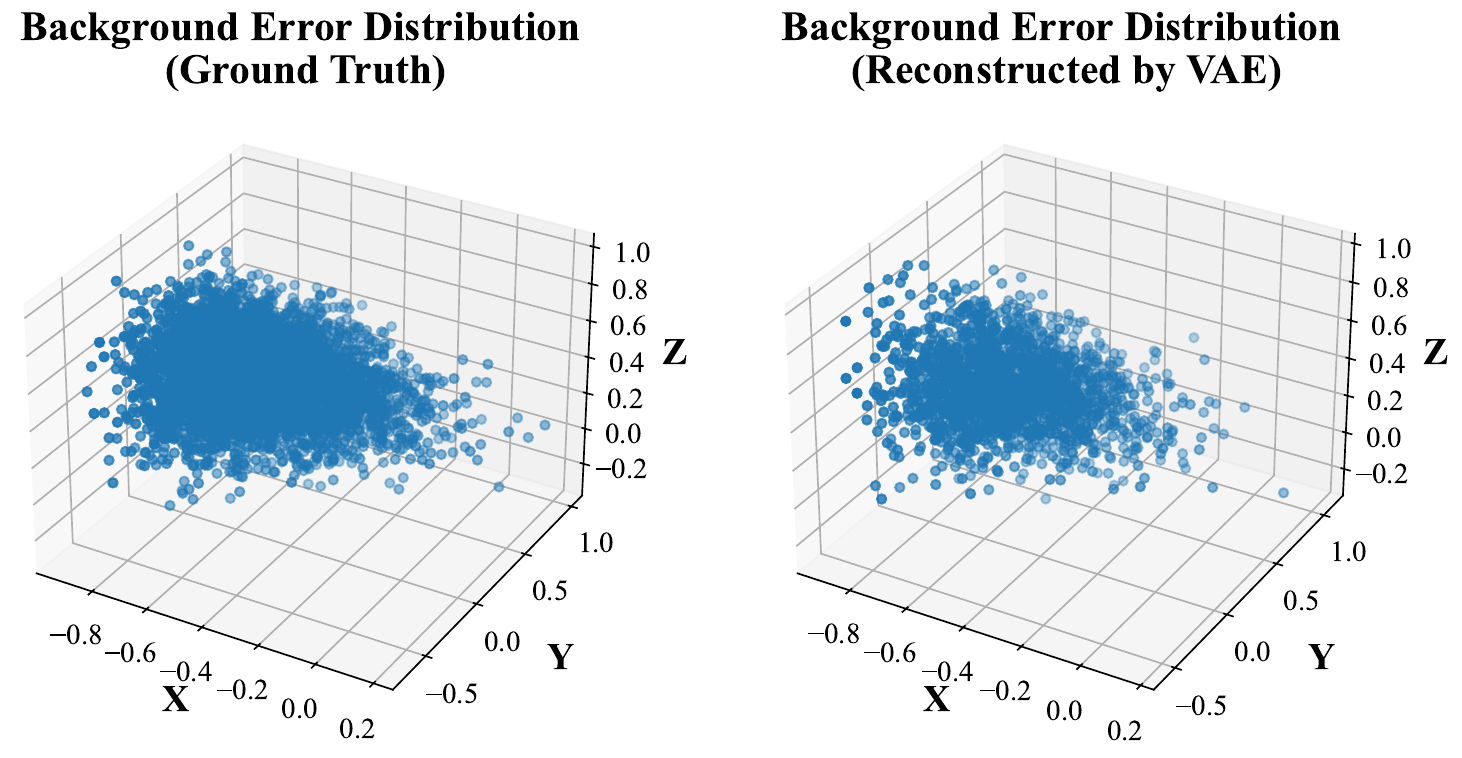}
    \caption{Background error distribution of the first three variables for the Lorenz 96 system when the ODE parameter is changed from $F_1=8$ to $F_1=13$. The left panel shows the ground truth background error distribution, and the right panel shows the background error distribution reconstructed by our trained VAE decoder. }
    \label{fig-bg-lorenz96-F}
\end{figure}

In the ODE of the Lorenz 96 system, there are totally $d$ equations. In each equation, there is one parameter to be tuned. Since all of the $d$ equations are symmetric, we only perturb the parameter of the first equation to construct the prediction model. The background error distributions (the first three variables) of both the ground truth and those constructed by our trained VAE are shown in Figure~\ref{fig-bg-lorenz96-F}. 

\subsubsection{Linear Observation Operator}

\begin{figure}[H]
    \centering
\includegraphics[width=1.0\linewidth]{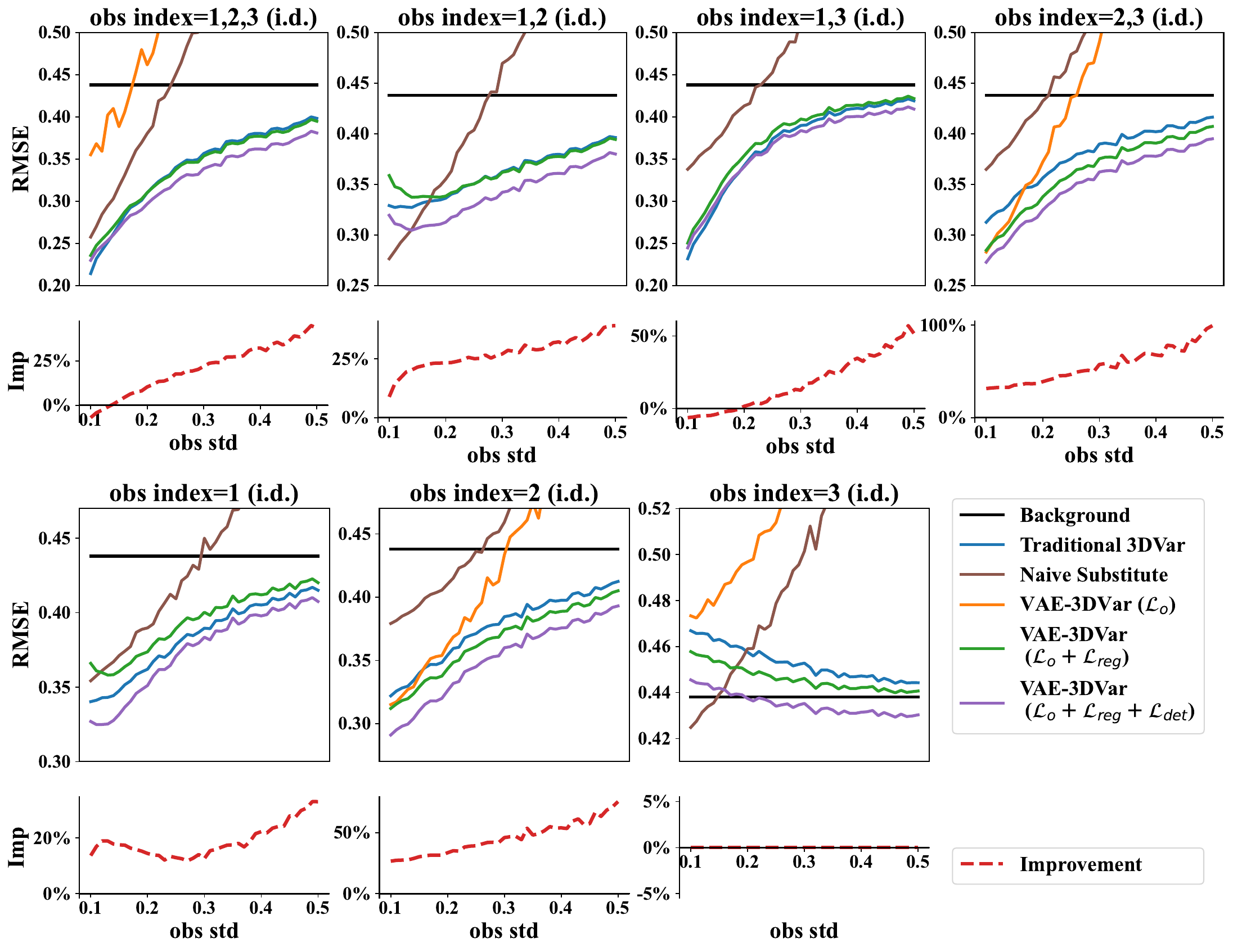}
    \caption{Results on the Lorenz 96 system with linear observation operators under 3DVar observational settings. The background states are constructed by changing the ODE parameter from $F_1=8$ to $F_1=13$. The metric $\mathrm{Imp}$ is set to zero if the analysis state after assimilation is worse than the background state.}
    \label{fig-lorenz96-3dvar-linear-appendix}
\end{figure}

We design seven different observation masks for evaluation on the Lorenz 96 system, iterating through all the possibilities of observing the first three variables. The results are reported in Figure~\ref{fig-lorenz96-3dvar-linear-appendix}. It can be found that in the Lorenz 96 system, VAE-3DVar almost outperforms the traditional 3DVar in all cases, except in cases when the observation noise is so small that an accurate estimate of background errors does not matter.

\subsubsection{Non-linear Observation Operator}

\begin{figure}[H]
    \centering
\includegraphics[width=1.0\linewidth]{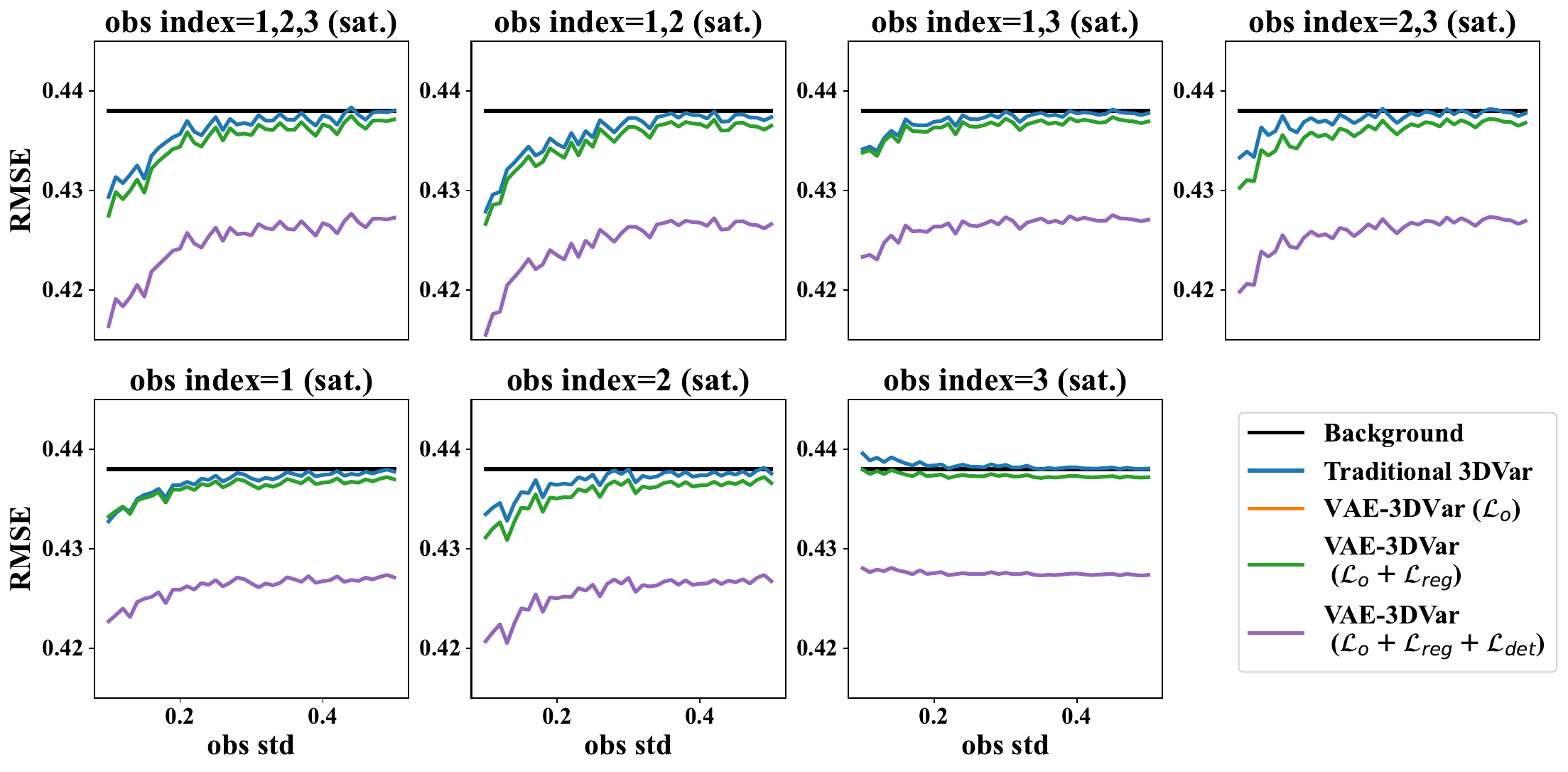}
    \caption{Results on the Lorenz 96 system with nonlinear observation operators under 3DVar observational settings. The observation operator is the saturated function. The background states are constructed by changing the ODE parameter from $F_1=8$ to $F_1=13$. }
    \label{fig-lorenz96-3dvar-nonlinear-appendix}
\end{figure}

In this section, the saturated function is applied to construct a non-linear observation operator. As shown in Figure~\ref{fig-lorenz96-3dvar-nonlinear-appendix} , we find a substantial improvement in accuracy with VAE-3DVar over traditional 3DVar. This is because the saturated operator reduces the magnitude of observations and introduces increased noise, thereby emphasizing the critical need for more precise estimation of the background error.

\subsection{Lorenz 96 (4DVar Settings)}

\begin{figure}[H]
    \centering
\includegraphics[width=1.0\linewidth]{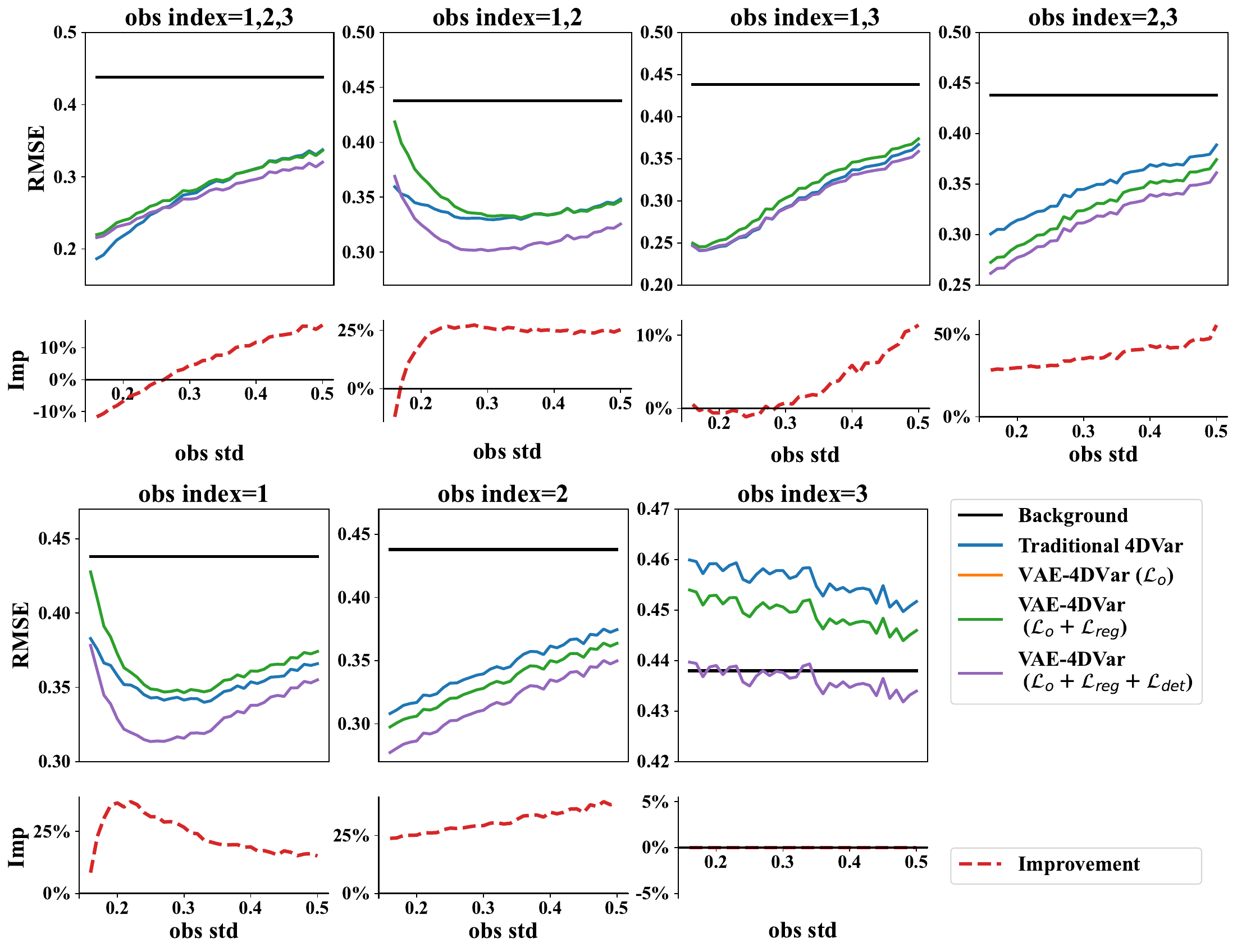}
    \caption{Results on the Lorenz 96 system with linear observation operators under 4DVar observational settings. The background states are constructed by changing the ODE parameter from $F_1=8$ to $F_1=13$. The metric $\mathrm{Imp}$ is set to zero if the analysis state after assimilation is worse than the background state. }
    \label{fig-lorenz96-4dvar-appendix}
\end{figure}

The 4DVar observational settings are also included for evaluation with a linear observation operator. The results, as shown in Figure~\ref{fig-lorenz96-4dvar-appendix}, are similar to that of the 3DVar setting, with a maximum improvement of over 50\%.

\begin{table}[H]
\centering
\caption{List of figures for additional results.  } 
\centering
\begin{tabular}{ccccc}
\toprule
    Figure index & Dynamical system & Parameters perturbed & 3DVar/4DVar & Observation Operator\\
\midrule
    \ref{fig-lorenz63-3dvar-linear-appendix} & Lorenz 63 & $\sigma$ & 3DVar & linear \\ 
    \ref{fig-lorenz63-3dvar-rho-appendix} & Lorenz 63 & $\rho$ &  3DVar & linear \\ 
    \ref{fig-lorenz63-3dvar-nonlinear-appendix} & Lorenz 63 & $\sigma$ & 3DVar & non-linear \\ 
    \ref{fig-lorenz63-4dvar-appendix} & Lorenz 63 & $\sigma$ & 4DVar & linear \\ 
    \ref{fig-lorenz96-3dvar-linear-appendix}  & Lorenz 96 & $F$ & 3DVar & linear \\ 
    \ref{fig-lorenz96-3dvar-nonlinear-appendix} & Lorenz 96 & $F$ & 3DVar & non-linear \\ 
    \ref{fig-lorenz96-4dvar-appendix} & Lorenz 96 & $F$ & 4DVar & linear \\ 
\bottomrule
\end{tabular}
\end{table}

\newpage

\newpage

\newpage

\newpage

\newpage

\newpage

\end{document}